\documentclass[10pt,letterpaper]{article}
\usepackage[left=1.7in]{geometry}

\usepackage{changepage}

\usepackage[utf8]{inputenc}

\usepackage{textcomp,marvosym}

\usepackage{fixltx2e}

\usepackage{amsmath,amssymb}

\usepackage{cite}

\usepackage{nameref,hyperref}

\usepackage[right]{lineno}

\usepackage{microtype}
\DisableLigatures[f]{encoding = *, family = * }

\usepackage{rotating}

\usepackage{pdfpages}

\usepackage[aboveskip=1pt,labelfont=bf,labelsep=period,singlelinecheck=off]{caption}

\makeatletter
\renewcommand{\@biblabel}[1]{\quad#1.}
\makeatother

\date{}

\usepackage{lastpage,fancyhdr,graphicx}
\usepackage{epstopdf}
\usepackage{graphicx}
\usepackage{url}
\usepackage[numbers,sort]{natbib}
\usepackage{todonotes}
\usepackage{multirow}
\usepackage{booktabs}
\usepackage[titletoc,title]{appendix}
\usepackage{tablefootnote}
\usepackage{placeins}

\begin{document}

\begin{flushleft}
{\Large
\textbf\newline{Evolution of Collective Behaviors for a Real Swarm of Aquatic Surface Robots}
}
\newline
\\
Miguel Duarte\textsuperscript{1,2,3,*},
Vasco Costa\textsuperscript{1,2,3},
Jorge Gomes\textsuperscript{1,2,4},
Tiago Rodrigues\textsuperscript{1,2,3}, \\
Fernando Silva\textsuperscript{1,2,4},
Sancho Moura Oliveira\textsuperscript{1,2,3},
Anders Lyhne Christensen\textsuperscript{1,2,3}
\\
\bigskip
\bf{1} BioMachines Lab, Lisbon, Portugal
\\
\bf{2} Instituto de Telecomunicações, Lisbon, Portugal
\\
\bf{3} Instituto Universitário de Lisboa (ISCTE-IUL), Lisbon, Portugal
\\
\bf{4} BioISI, Faculdade de Ciências, Lisbon, Portugal
\\
\bigskip

%
%





* miguel\_duarte@iscte.pt

\end{flushleft}


\section*{Abstract}

Swarm robotics is a promising approach for the coordination of large numbers of robots. While previous studies have shown that evolutionary robotics techniques can be applied to obtain robust and efficient self-organized behaviors for robot swarms, most studies have been conducted in simulation, and the few that have been conducted on real robots have been confined to laboratory environments. In this paper, we demonstrate for the first time a swarm robotics system with evolved control successfully operating in a real and uncontrolled environment. We evolve neural network-based controllers in simulation for canonical swarm robotics tasks, namely homing, dispersion, clustering, and monitoring. We then assess the performance of the controllers on a real swarm of up to ten aquatic surface robots. Our results show that the evolved controllers transfer successfully to real robots and achieve a performance similar to the performance obtained in simulation. We validate that the evolved controllers display key properties of swarm intelligence-based control, namely scalability, flexibility, and robustness on the real swarm. We conclude with a proof-of-concept experiment in which the swarm performs a complete environmental monitoring task by combining multiple evolved controllers.


\section{Introduction}
\label{sec:introduction}

Swarm robotics systems (SRS) are a promising approach to collective robotics, in which large groups of relatively simple and autonomous robots are able to display collectively intelligent behavior~\cite{brambilla2013swarm}. Control in a SRS is decentralized, meaning that each individual robot operates based on its local observations of the environment and coordination with neighboring robots. During task execution, the swarm-level behavior emerges from the interactions between neighboring robots, and from the interactions between robots and the environment. The swarm robotics approach has the potential to incite a number of important properties in multirobot systems~\cite{brambilla2013swarm,bayindir2007review,Bayindir2015}, namely: (i)~robustness, the ability to cope with the faults of individual robots; (ii)~flexibility, the ability to operate in a variety of different environments and to perform various tasks; and (iii)~scalability, the ability to generally maintain the group behavior regardless of the size of the swarm. Due to these properties, SRS have an enormous potential in several real-world domains, such as search and rescue, exploration, surveillance, and clean up~\cite{bayindir2007review,brambilla2013swarm}. Despite the potential of swarm robotics for real-world tasks, no demonstrations of swarm behaviors outside of controlled laboratory conditions have been performed so far. As discussed by Brambilla~\emph{et al.}~\cite{brambilla2013swarm} and Bay{\i}nd{\i}r~\cite{Bayindir2015} in recent reviews of the field of swarm robotics, all real-robot experiments presented in the literature covered have, in fact, been performed in controlled laboratory environments such as enclosed arenas in which the relevant conditions are defined by the experimenter. That is, although SRS are ultimately intended to operate in potentially unstructured, real-world environments, experiments have been conducted mainly in simulation, and at the current state of development no study has been able to demonstrate the benefits of SRS in real-world environments.

One of the key challenges in SRS is the synthesis of behavioral control. Manual design of control for each robot requires the decomposition of the macroscopic, swarm-level behavior into microscopic behavioral rules that determine the interactions between neighboring robots, and between the robots and the environment so that the global, collectively self-organized behavior emerges. The designer thus needs to be able to understand the relation between local robot interactions and emerging swarm-level properties. There is, however, no approach to derive the microscopic behavioral rules based on a desired global behavior or task description for the general case~\cite{dorigo2004evolving}. In this respect, the use of evolutionary computation techniques has been studied as a promising alternative to traditional control approaches, due to its capability to automatically synthesize self-organized control based only on the definition of a high-level performance metric (see~\cite{nolfi2000evolutionary,bongard2013evolutionary,lipson2000automaticrobot} for examples).

SRS are typically associated with ground robots~\cite{Dorigo13,brambilla2013swarm} and, more recently, with aerial robots~\cite{lindsey2012construction,basiri2014audio}. In this paper, we show that SRS also have the potential to be applied to maritime tasks, which are usually expensive to carry out because they rely on manned vehicles with large operational crews. While significant effort has been made to adapt unmanned vehicle technology to maritime tasks, such systems are currently relatively expensive to acquire and operate, and only a single or a few robots are typically deployed~\cite{yan2010development}. 
The use of SRS at sea is advantageous because: (i)~several maritime tasks, such as environmental monitoring, sea-life localization, and sea-border patrolling, require distributed sensing at a high spatial and temporal resolution, and are therefore challenging to carry out with only a single or a few robots; and (ii)~robots operating at sea need to display a high degree of fault-tolerance and robustness~\cite{yan2010development}, which can be provided by SRS approaches~\cite{sahin2005}.

\begin{figure}[h!]
\centering
\includegraphics[width=\textwidth]{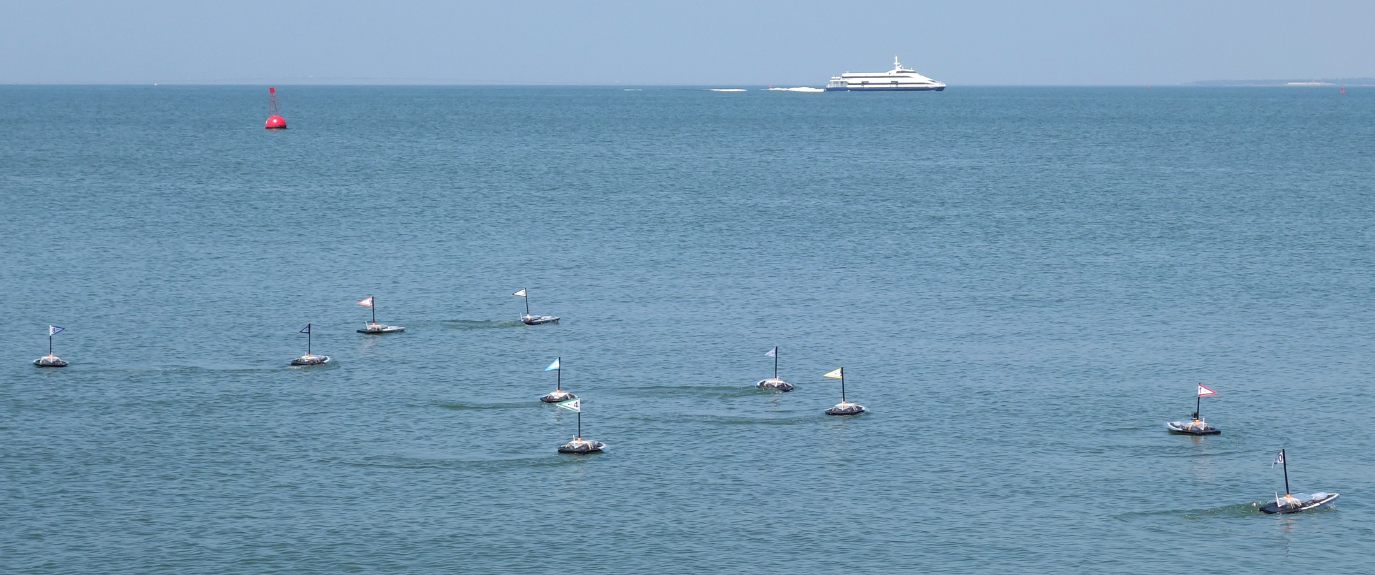}
\caption{The robotic swarm with 10 units performing a homing task.}
\label{fig:alldrones}
\end{figure}

In this paper, we show for the first time a SRS system with evolved, distributed control performing self-organized behaviors in real-world environments. Our study relies on a swarm of up to ten small, simple, and inexpensive aquatic surface robots (see Fig.~\ref{fig:alldrones}). First, we demonstrate our SRS performing four canonical tasks in an uncontrolled aquatic environment: (i)~homing, (ii)~dispersion, (iii)~clustering, and (iv)~monitoring. Second, we perform a set of experiments to demonstrate the scalability and robustness of the evolved controllers when operating in a real environment. Finally, we show how multiple evolved swarm behaviors can be combined to produce more sophisticated controllers, which we use to perform an environmental monitoring task where the robots have to sample the water temperature in a given area. In summary, our experiments demonstrate that a swarm of robots with evolved control can operate successfully in a real-world environment.

\section{Related work}
\label{sec:related_work}



\subsection{Swarm Robotics}

The field of swarm robotics, as well as the more general field of swarm intelligence, takes inspiration from the observation of social insects, such as ants, bees, wasps, and termites~\cite{sahin2005}. In these animal societies, relatively simple units rely on self-organization to display collectively intelligent behavior. The self-organized behavior of social insects confers the swarms a high level of scalability, versatility, and robustness to individuals failures~\cite{camazine03}. The key motivation behind swarm robotics is to harness these properties to build large-scale decentralized multirobot systems~\cite{sahin2005,jones2005behavior}. Due to their properties, SRS have significant potential to take on a number of applications requiring distributed sensing and/or action. In a recent review, Bayindir~\cite{Bayindir2015} discusses the categories of behaviors that have been achieved with SRS, such as aggregation, flocking, foraging, object clustering and sorting, navigation, path formation, swarm self-deployment, collaborative manipulation, and task allocation.

Despite the advances in the field, SRS have, however, up to this point not leveraged their full potential, and have been confined to controlled laboratory environments~\cite{brambilla2013swarm}, still far from real-world applications. There are multiple reasons for the absence of real-world SRS, one of the most prevalent being the difficulty in designing behavioral control for the individual robots that results in the desired swarm-level behavior. This problem is exacerbated when trying to achieve behaviors capable of dealing with the complexity of real tasks and uncontrolled environments, as opposed to the abstract tasks and highly controlled environments that have been used in the majority of previous works~\cite{brambilla2013swarm,Bayindir2015}.

\subsection{Evolutionary Synthesis of Control for SRS}

Evolutionary robotics (ER)~\cite{nolfi2000evolutionary} is the research field that studies the application of evolutionary computation to the synthesis of robotic control. ER has become a popular alternative to manual programming in a variety of SRS applications~\cite{brambilla2013swarm}. Given the specification of a task, an evolutionary algorithm evaluates and optimizes controllers in a holistic manner, thereby facilitating the emergence of self-organized behavior~\cite{nolfi98selforganization,trianni2011engineering}. That is, based on the performance of the resulting swarm-level behavior, the evolutionary algorithm iteratively fine-tunes the parameters and the microscopic rules governing each individual robot, thus eliminating the need for manual and detailed specification of low-level control~\citep{FloreanoKeller2010}. Optimization of genomes is based on Darwinian evolution, namely blind variations and survival of the fittest.

Evolutionary approaches have been used to solve a large number of swarm robotics tasks such as coordinated motion~\cite{baldassarre07,sperati2008evolving}, chain formation~\cite{sperati2011self}, aggregation~\cite{trianni2003evolving,soysal2007aggregation,gomes2013noveltyswarms}, flocking~\cite{baldassarre2003evolving}, hole avoidance~\cite{trianni2006cooperative}, aerial vehicles communication \cite{hauert09aerial}, categorization~\cite{ampatzis08}, group transport~\cite{grob2008evolution,gross2009grouptransport}, social learning \citep{pini08}, sharing of an energy recharging station~\cite{gomes2013noveltyswarms,gomes13gecco}, and patrolling of an area~\cite{duarte2014aquatic}. The majority of these studies were, however, performed exclusively in a simulated environment. The limited number of studies that used real robots have been conducted mostly with small (up to 15\,cm) wheeled robots~\cite{khaldi2015overview}, such as the \emph{Khepera}~\cite{mondada1999development}, the \emph{s-bot}~\cite{mondada2004swarm} and the \emph{e-puck}~\cite{mondada2009puck}. These studies relied on abstract tasks in highly controlled environments: relatively small and enclosed arenas in laboratories.


The use of simulation in ER is widespread due to the large number of evaluations required by an evolutionary process to find adequate controllers. Transferring controllers evolved in simulation to real robots is currently one of the biggest challenges in ER~\cite{doncieux2015evolutionary}, since the evolutionary process tends to exploit simulation-specific features that are not present in the real world, resulting in the real robots behaving differently and obtaining a lower task performance than in simulation. This issue is usually referred to as the \emph{reality gap}~\cite{Jakobi97evolutionaryrobotics}.

Several approaches have been proposed to try to overcome the reality gap. Miglino et al.~\cite{Miglino96evolvingmobile} suggested three complementary techniques: (i)~using samples from the real robots' sensors to more accurately model them in simulation, (ii)~introducing a conservative form of noise to promote the evolution of robust controllers, and (iii)~continuing evolution in real hardware to tune controllers to the differences between simulation and reality. Jakobi~\cite{Jakobi97evolutionaryrobotics} proposed the use of \emph{minimal simulations}, in which the simulation model is based on a small set of features deemed necessary for successful synthesis of controllers, while remaining features are masked by an envelope of noise. Koos et al.~\cite{koos2013transferability} proposed \emph{the transferability approach}, a multi-objective formulation of ER in which controllers are evaluated both by their performance in simulation and their performance on real robots. The approach requires periodical transfer of control during the evolutionary process in order to update a surrogate model, which estimates how well candidate solutions will transfer to the real world. Lehman et al.~\cite{lehman2013encouraging} proposed a different approach where the evolutionary process explicitly rewards the evolved behaviors for displaying a high reactivity, as measured by the mutual information between the magnitude of changes in a robot's sensors and actuators. It is shown that controllers with a reactive disposition transfer better to real robots than those not rewarded for reactive behavior, regardless of the noise conditions with which the controllers were evolved.

Another challenge in evolutionary robotics is how to drive the evolutionary process towards high-quality solutions, avoiding potential local optima. When the task to which solutions are sought reaches a certain level of complexity, traditional evolutionary approaches are prone to suffer from bootstrap problems and deception~\cite{doncieux2014,silva_open_issues}. One approach to deal with these issues is to employ diversity maintenance techniques, such as novelty search~\cite{lehman2011abandoning}, in order to allow an evolutionary algorithm to explore multiple different paths through the search space~\cite{mouret2012encouraging,gomes2013noveltyswarms,gomes16ec}. A different approach is to directly assist the evolutionary process with domain-specific knowledge, by using techniques such as incremental evolution~\cite{mouret2008incremental,christensen2006incremental}, behavioral decomposition~\cite{moioli2008towards,lee1999evolving,duarte2015hybrid}, or semi-interactive human-in-the-loop-techniques~\cite{bongard2012avoiding,celis2013avoiding,woolley2014naiec}. In Section~\ref{sec:multicontroller_mission}, we demonstrate a simple form of behavioral decomposition: we show how different evolved behaviors can be combined to produce a more complex controller, capable of performing an entire mission.

\subsection{Aquatic Robots}

In recent years, significant development has been conducted in the areas of autonomous underwater vehicles and autonomous surface vehicles~\citep{manley2008unmanned}, focusing on individual robots with a high degree of hardware and software complexity. Such systems have been applied to a variety of scenarios, such as environmental monitoring~\cite{pinto2014robio} and search and rescue~\cite{de2013eu}. Nonetheless, they are expensive, and restricted in terms of the tasks they can undertake as they have a  limited degree of autonomy, such as basic waypoint navigation capabilities. There have been numerous conceptual studies on systems composed of multiple robots with decentralized control for aquatic environments. However, only few such systems have been realized. Scerri~\emph{et al.}~\cite{scerri2012real}, for instance,  made use of groups of up to five autonomous robots for distributed environmental monitoring missions. Their robots, however, need to be controlled by a central computer, which limits deployment in remote locations, and reduces their potential to scale to larger groups. Other studies in the domain of swarms of underwater robots have been performed in the scope of the CoCoRo project~\citep{schmickl2011cocoro}. Regarding the evolution of control for aquatic robots, significantly fewer studies have approached the subject. Examples include the evolution of station keeping abilities for a single, simulated underwater robot~\citep{moore2013evolution}, and the evolution of \emph{centralized} control for a team of four underwater robots in a simulated predator-prey task~\citep{praczyk2014augmenting}.


\section{Methodology}
\label{sec:methodology}

\begin{figure}[h]
	\centering
	\includegraphics[width=\linewidth]{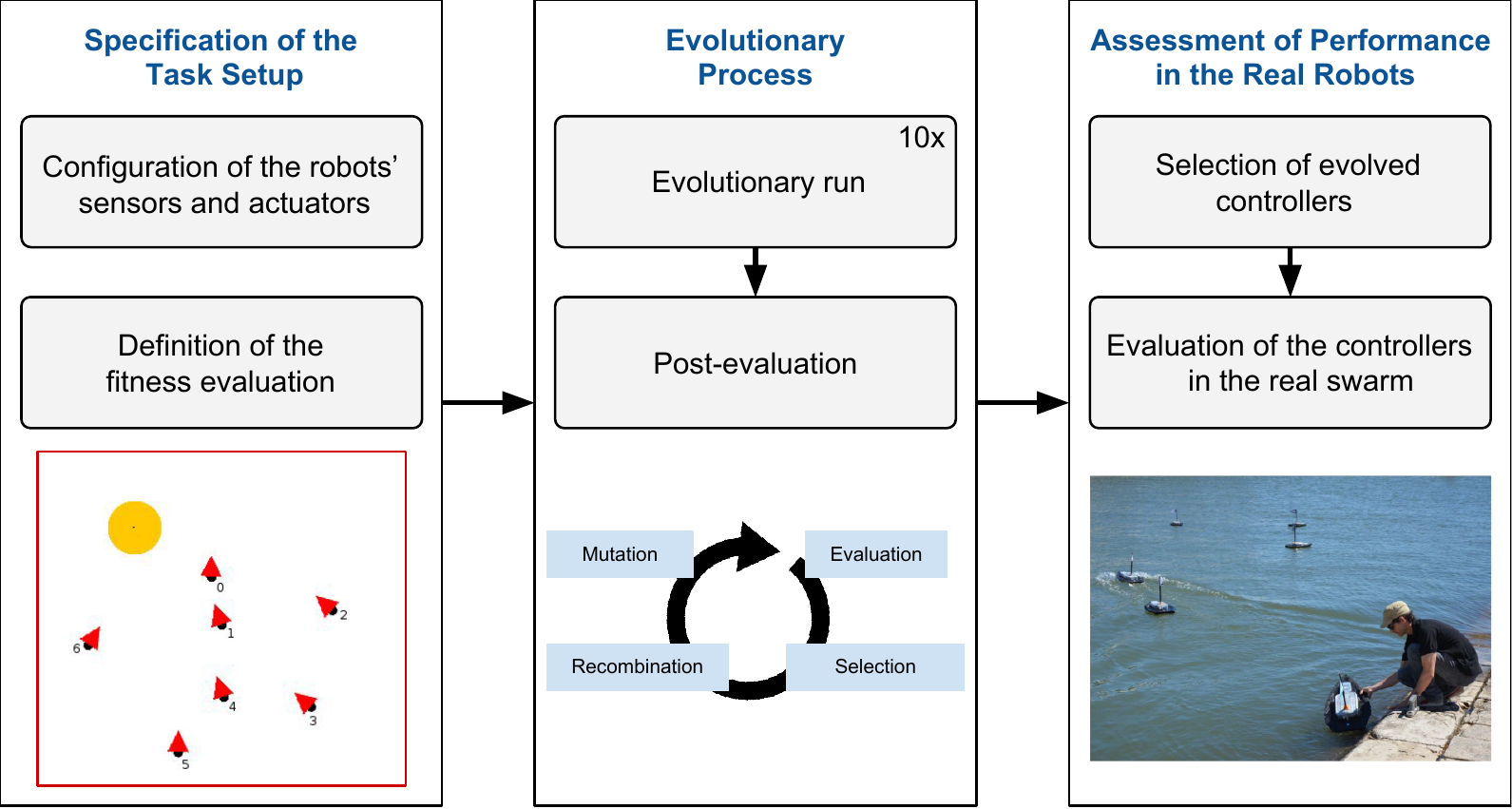}
	\caption{The control synthesis and performance assessment process for each task.}
	\label{fig:evolution_flow}
\end{figure}

In this section, we present the methodology that we followed to produce controllers for the four tasks, and the steps we took to ensure a successful transfer of the controllers to the real robotic swarm. The methodology is summarised in Fig.~\ref{fig:evolution_flow}.

\paragraph{1. Definition of the simulation platform}

The evolutionary process used to synthesize behavior for the swarm relied exclusively on simulation for the evaluation of the candidate solutions. It is therefore necessary to implement a simulator in which the performance of the controllers can be assessed. We used a two-dimensional simulation environment, where the robots were abstracted as circular objects with a certain heading and position. The general principle behind the simulation was to model the motion of the robots based on real measurements taken in the water, but without including physics simulation and fluid dynamics, which would have resulted in a complex and computationally expensive simulation environment. Maintaining a low computational complexity is essential for making the evolutionary robotics process feasible, as a large number of simulations have to be conducted for each evolutionary run.

We developed a middle-layer that is shared by both the on-board software and the simulation platform. This enabled the same code-base and controllers to be executed on both the real robots and on robots in simulation, thereby minimizing the implementation differences between simulation and reality.

\paragraph{2. Configuration of the robots' sensors and actuators}

The processing of raw sensory data into values that can be fed to the neural controller is an essential step to enable the evolution of control. To this end, the robots (both real and simulated) relied on sensors that return the distance to objects of interest in equal-sized segments around the robot.  These sensor values are obtained by pre-processing the GPS locations of entities in the environment of which the robot is currently aware, and the current heading and position of the robot.  It should be noted that the sensors and actuators used by the real robots are inherently associated with a high degree of stochasticity. To account for these phenomena in simulation, we added noise to the sensors and actuators of the simulated robots, based on the observations made on the real robots. This conservative noise approach~\cite{Miglino96evolvingmobile} is simple to implement, requires only basic assumptions and domain knowledge, and has proven effective in previous studies conducted in controlled conditions in our lab~\cite{duarte2015hybrid}. The noise included (i)~the variation of robot parameters at the beginning of each simulation trial (e.g. maximum motor speed), (ii)~noise injected at every simulation step (e.g. errors in sensory readings), and (iii)~environmental noise (water currents).

\paragraph{3. Definition of the fitness evaluation}

For each task, we specified a fitness function that translated the task objective that needed to be achieved. The task setup chosen for the evaluation of controllers for each task approximated the setup in which the real robots were going to be used, meaning that the environment had approximately the same size, the number of robots was within the same range, and the robots had the same capabilities. To evaluate each candidate solution (controller), 10 independent simulation trials were conducted, and the fitness of the controller was the mean score obtained in these simulations. To avoid overfiting the controllers to specific task conditions, we introduced random variations in the simulation trials during the evolutionary process. In every trial, we varied the number of robots (between five and ten), their starting position and orientation, and the location of other spatial entities (when applicable).

\paragraph{4. Evolutionary process}

The evolutionary algorithm optimized the neural network-based controllers for the robots. We used the NEAT algorithm~\citep{stanley2002evolving} to evolve the neural network controllers for all tasks. NEAT is a widely used neuroevolution algorithm that simultaneously optimizes the connection weights and the topology of artificial neural networks. The default NEAT parameters were were used for all different tasks studied in this paper. Each evolutionary setup was repeated for 10 independent evolutionary runs. After each evolutionary run, we post-evaluated the top controller of each generation in 100 simulations to obtain an accurate estimate of the controllers' quality.

\paragraph{5. Selection of evolved controllers}

For each task, we identified the highest-performing controller of each evolutionary run based on the post-evaluation results. Out of these ten best-of-run controllers, we then selected the three with the highest fitness scores, as separate evolutionary runs are likely to produce different controllers.

\paragraph{6. Assessment of performance in the real robots}

We finally assessed the real-robot performance of the top three controllers for each task, and compared their performance to that obtained in simulation. Each selected controller was evaluated in three independent experiments. For every experiment, the initial positions of the robots were randomized: a set of positions was generated inside a task-specific region, and each robot navigated to its assigned starting position. When all robots were in place, the experiment started and the controllers were simultaneously activated in all robots. The experiment ended after a fixed amount of time. When evaluating the three different controllers for a given task, the same set of initial positions was used. To establish a fair comparison with controller performance in simulation, the same set of initial conditions were also used, i.e., the initial positions logged by robots were used as the initial robot positions for the simulation trials.

\section{Experimental Setup}
\label{sec:experimental_setup}


\subsection{Robotic Platform}
\label{sec:hardware}

\begin{figure}[h!]
\centering
\includegraphics[width=.6\linewidth]{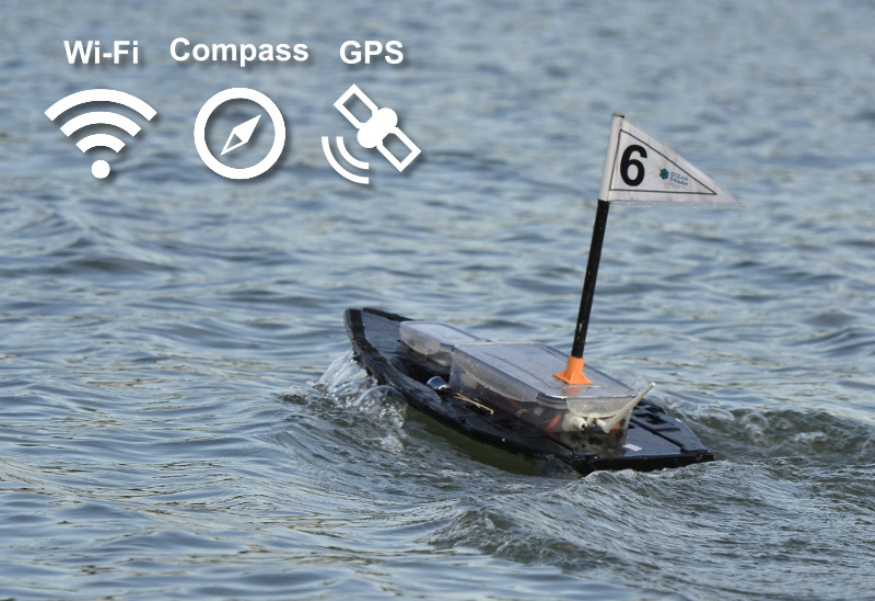}
\caption{The robot is an autonomous surface vehicle equipped with Wi-Fi for communication, and a compass and GPS for navigation. It has a length of 60\,cm and can move at speeds of up to 1.7\,m/s.}
\label{fig:robotcloseup}
\end{figure}

We designed and produced a total of 10 simple, small (60~cm in length) and inexpensive ($\approx$ 300~eur/unit) robots, using widely available, off-the-shelf hardware. Each robot model is a differential drive mono-hull boat (see Fig.~\ref{fig:robotcloseup}). The maximum speed of the robot is 1.7\,m/s (3.3\,kts), and the maximum angular velocity is 90$^\circ$/s. The on-board control of each robot was executed by a Raspberry Pi 2 single-board computer. The robots form a distributed network without any central coordination or single point of failure through a IEEE 802.11g (Wi-Fi) ad-hoc wireless network, and communicate with one another by broadcasting UDP datagrams. Each robot is equipped with a GPS and a compass, and broadcasts its position to neighboring robots every second. See Supporting Information S1 Text for additional information on the hardware configuration of the robots.

The robotics simulation was performed on JBotEvolver~\citep{duarte2014jbotevolver} -- an open-source framework that has been used in a large number of evolutionary robotics studies, and is available at \url{https://github.com/BioMachinesLab/jbotevolver} under GNU GPL license. The robot's dynamic model was implemented based on measurements taken from a single robot. We measured the robot's movement with multiple combinations of motor speeds, which allowed us to characterize the motion dynamics, friction, and inertia. We then tuned the model's parameters by comparing simulated and real motion patterns.



\begin{figure}[t!]
	\centering
	\includegraphics[width=.85\textwidth]{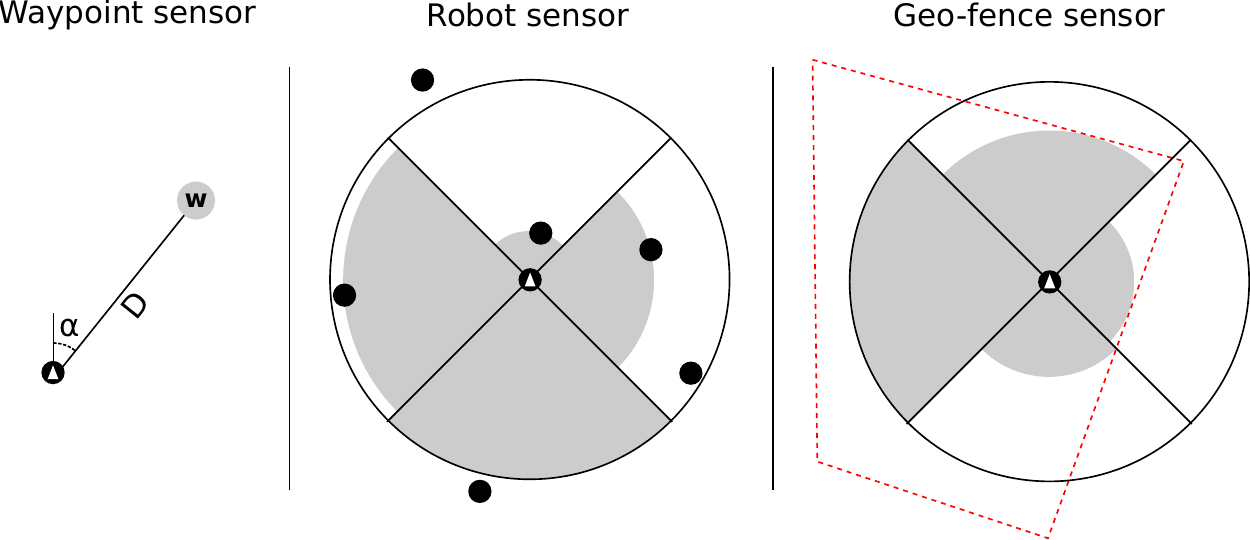}
	\caption{Illustration of the three types of sensors.}
	\label{fig:sensors}
\end{figure}

Each robot was controlled by an artificial neural network. The inputs of the neural network were the normalized readings of the sensors, and the outputs of the network controlled the robot's actuators. The sensor readings and actuation values were updated every 100\,ms. The neural network controlling each robot had two output neurons which controlled the linear speed and the angular velocity of the robot. These two outputs were then converted to left and right speeds and applied to the robot's motors. We implemented three sensors for the detection of points and objects of interest in the task environment, based on GPS locations and on information shared by neighboring robots (see Fig.~\ref{fig:sensors}).

\begin{description}
\item[Waypoint sensor:] This sensor is used to locate waypoints in the environment, if any. The sensor returns two values: (i)~the relative angle from the robot to the waypoint, and (ii)~the distance from the robot to the waypoint.
\item[Robot sensor:] This sensor measures the distance to nearby robots. The area around the robot is divided into four equally-sized slices. The sensor returns four values, one per slice, linearly proportional to the distance between the robot and the closest neighbor in that slice, or the maximum value if there is none within the sensor range.
\item[Geo-fence sensor:] This sensor is used to detect the geo-fence. Similarly to the robot sensor, the geo-fence sensor returns four values, one for each slice around the robot, indicating the minimum distance to the fence, or the maximum sensor value if the fence is outside the sensing range. One additional value is returned, indicating whether the robot is inside the fence or not (the geo-fence is a polygon).
\end{description}



\subsection{Evolutionary Setup and Task Configurations}
\label{sec:evolutionary_setup}


In this paper, we study the evolution of controllers for a real swarm of aquatic robots operating in a large uncontrolled environment. Our study relies on four cooperative tasks, which have been previously studied in simulation and laboratory environments~\cite{Bayindir2015}: (i)~homing in group while avoiding collisions, (ii)~dispersion, (iii)~clustering, and (iv)~monitoring of a pre-defined area. In this section, we describe the evolutionary setup that was used to synthesize control for each task, including the fitness functions and relevant parameters. All evolutionary and task parameters are listed in Supporting Information S1 Text.


\subsubsection*{Homing}

Navigation and obstacle avoidance is an essential feature for autonomous robots operating in the real world. This type of tasks was among the first to be studied in the field of evolutionary robotics~\cite{floreano1996evolution}. A variation of the task, known as \textit{homing}, involves one or more robots moving towards a point of interest in the environment, sometimes with obstacles in the way~\cite{christensen2006evolving}. 

In the homing task used in our experiments, the swarm of robots had to navigate to a given waypoint while avoiding collisions between the robots. To evolve solutions for this task, we rewarded controllers for minimizing the average distance of the swarm to a waypoint, according to the following equation:

\begin{equation}
f_{\mathit{homing}} = \left(\frac{1}{T}\sum_{t=1}^{T}{\frac{1}{R}\sum_{r=1}^{R}{\frac{\mathit{startingDist}_r - \mathit{dist}_{r,t}}{\mathit{startingDist}_r}}}\right) \times S \enspace ,
\end{equation}
where  $T$ is the maximum number of time steps, $R$ is the number of robots, $startingDist_r$ is the initial distance from robot $r$ to the waypoint, and $\mathit{dist}_{r_t}$ is the distance of robot $r$ to the waypoint at time $t$.

Collisions between robots are undesirable in any task, as they can damage the real robots. To avoid collisions, we introduced a \emph{safety coefficient} ($S$) when assessing the fitness in all task setups, which penalized solutions where robots get too close to one another (less than 3\,m). The safety coefficient $S$ is in the range [0.1,1], and is inversely proportional to the minimum distance between any two robots in the current simulation trial ($minDist$). The safety coefficient is defined according to the following equation:
\begin{equation}
S = 0.1 + \frac{\mathit{max}(0,\mathit{min}(3,\mathit{minDist}))}{3} \times 0.9 \enspace ,
\end{equation}

\subsubsection*{Dispersion}

In a dispersion task, each individual robot in the swarm has to maintain a predefined distance (\emph{target distance}) from its nearest neighbor. Robots should cover a large area without risking losing contact with the rest of the swarm, which can be an issue in unbounded environments. Examples in the literature include solutions that rely on preprogrammed behaviors~\cite{batalin2002spreading}, potential fields~\cite{reif1999social}, automatically synthesized state-machines~\cite{francesca2014experiment}, and artificially evolved ANNs~\cite{duarte2014aquatic}.

For this task, we chose a target distance of 20~m (half of the robots' communication range). During evolution, controllers were rewarded for minimizing the difference between the current distance to the nearest neighbor and the target distance, according to the following equation:


\begin{equation}
f_{\mathit{dispersion}} = \left(\frac{1}{T}\sum_{t=1}^{T}{\frac{1}{R}\sum_{r=1}^{R}{1 - |\mathit{minDist}_{r,t} - \mathit{targetDist}|}}\right) \times S \enspace ,
\end{equation}
where  $T$ is the maximum number of time steps, $R$ is the number of robots, $\mathit{minDist}_{r_t}$ is the distance of robot $r$ to the nearest neighbor at time $t$, and $\mathit{targetDist}$ is the distance at which the robots should disperse.

\subsubsection*{Clustering}

Clustering, also known as \textit{aggregation}, is a canonical task in swarm robotics~\cite{brambilla2013swarm}. Clustering is a challenging task because it combines several aspects of multirobot tasks~\cite{silva2012odneat}, including distributed individual search, coordinated movement, and cooperation. Furthermore, clustering plays an important role in robotics because it is a precursor of other collective behaviors such as group transport of heavy objects~\cite{gross2009grouptransport}.

In our instance of the clustering task, the robots started randomly spread in a square-shaped area with a side-length of 100\,m, and had the objective of finding one another so as to form a single cluster. However, since the communication range was limited and the environment was unbounded, certain initial conditions could lead to the formation of more than one cluster. During evolution, controllers were scored based on the number of clusters that they formed. Two robots belonged to the same cluster if the distance between them was inferior to 7~m. The fitness function was defined based on a weighted sum of the number of clusters throughout time:


\begin{equation}
f_{\mathit{clustering}} = \frac{\sum_{t = 1}^{T}{ t \times \frac{R - c_t}{R - 1}}}{\sum_{t=1}^{T}{t}} \times S \enspace ,
\end{equation}
where $T$ is the total number of time steps in each trial, $R$ is the number of robots, and $c_t$ is the number of clusters formed by the robots at step $t$.

\subsubsection*{Area Monitoring}

Swarms of robots are ideal for tasks where large areas need to be covered for monitoring or surveillance purposes. This task category has been explored in the past with several different control techniques, such as pheromone traces~\cite{wagner1999distributed}, odor sources~\cite{marques2006particle} and evolved artificial neural networks~\cite{haasdijk2011racing}.

In our area monitoring task, a geo-fence is defined to delimit an area of interest, and the robots should coordinate to continuously cover as much of the area as possible. During evolution, we tested the controllers in a variety of randomly generated geo-fences with different shapes, in order to obtain general behaviors. For the purpose of this task, we considered that each robot covers a circular area with a radius of 5\,m ($V$) around it. The challenge in this task is to find a general movement pattern that takes into account many different shapes of the monitoring area and varying number of robots.

In order to assess the performance of the controllers, we divided the monitoring area into a grid with a fixed cell size of 1\,m$^2$. Each robot could visit cells within the coverage radius $V$, setting its value to 1. The value of previously visited cells decayed linearly over a time frame of 100\,s, down to 0. Controllers were scored based on how much of the grid was covered over time, according to the following equation:

\begin{equation}
\begin{split}
f_{\mathit{monitor}} &=  \left(\frac{1}{T}\sum_{t=1}^{T}\frac{1}{C} \sum_{c=1}^{C}{\mathit{val}(c_t)} \right) \times S \;,\\
val(c_t) &=
\begin{cases}
0 & ,\; t = 1 \\
\mathit{max}(0, \mathit{val}(c_{t-1}) - \Delta) & ,\; \mathit{minDist}_{c,t} > V \\
1 & ,\; \mathit{minDist}_{c,t} \leq V \\
\end{cases}
\end{split}
\end{equation}
where $T$ is the maximum number of time steps, $C$ is the number of cells, $\mathit{minDist}_{c,t}$ is the minimum distance of any robot to cell $c$ at time $t$, and $\Delta$ is the decay for any cell (0.001 per time step).

\subsection{Evolutionary Results}

\begin{figure}[h!]
\centering
\includegraphics{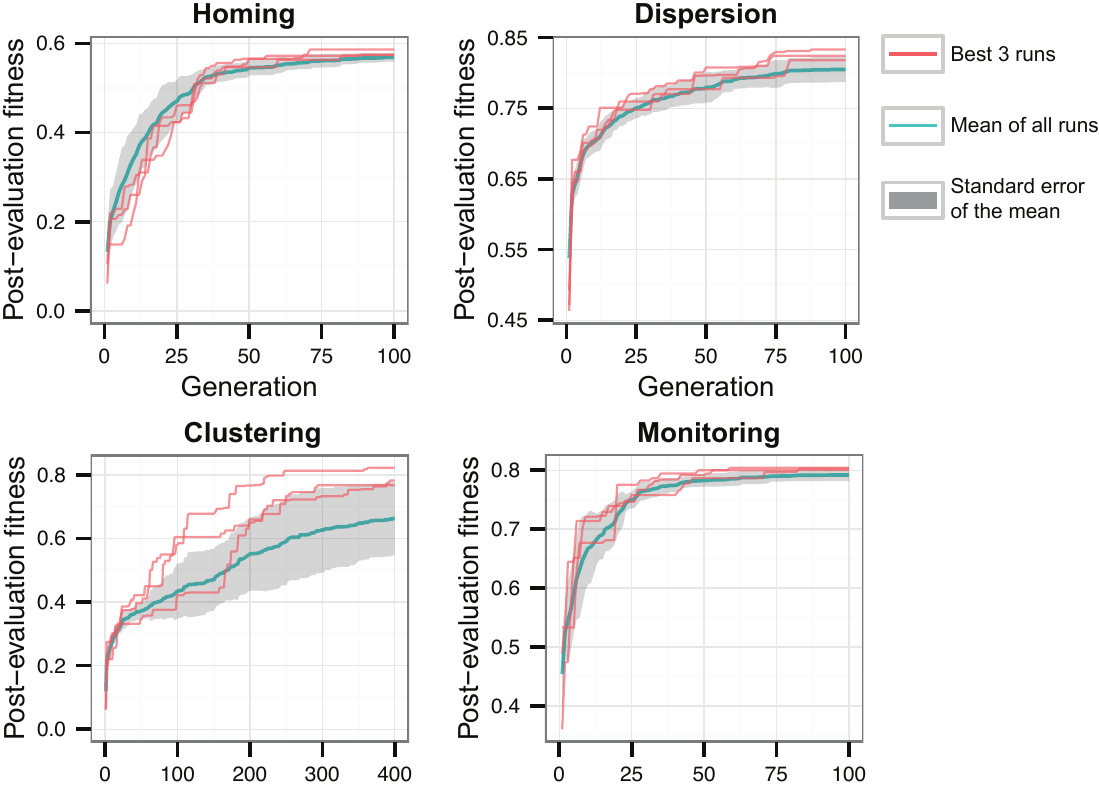}
\caption{Fitness plot for the four different tasks. The plot shows the highest fitness scores found so far at each generation. The red lines depict the three highest-scoring evolutionary runs, while the blue line depicts the average of the ten runs, with the respective standard deviation shown in gray.}
\label{fig:fitness}
\end{figure}

The results of the evolutionary algorithm for all tasks can be found in Fig.~\ref{fig:fitness}. Solutions for homing, dispersion, and monitoring were found within 100 generations, while clustering required up to 400 generations for effective solutions to evolve. This can be explained by the challenge in forming a single cluster from random initial positions: since the environment is unbounded, some robots might find themselves outside of each other's communication range, making the task more difficult. Section~\ref{sec:experiments_and_results} covers the results in detail, including comparisons between performance in simulation and on the real robots.

\section{Transferring Evolved Controllers to Real Robots}
\label{sec:experiments_and_results}

\begin{figure}[t!]
	\centering
	\includegraphics[width=\textwidth]{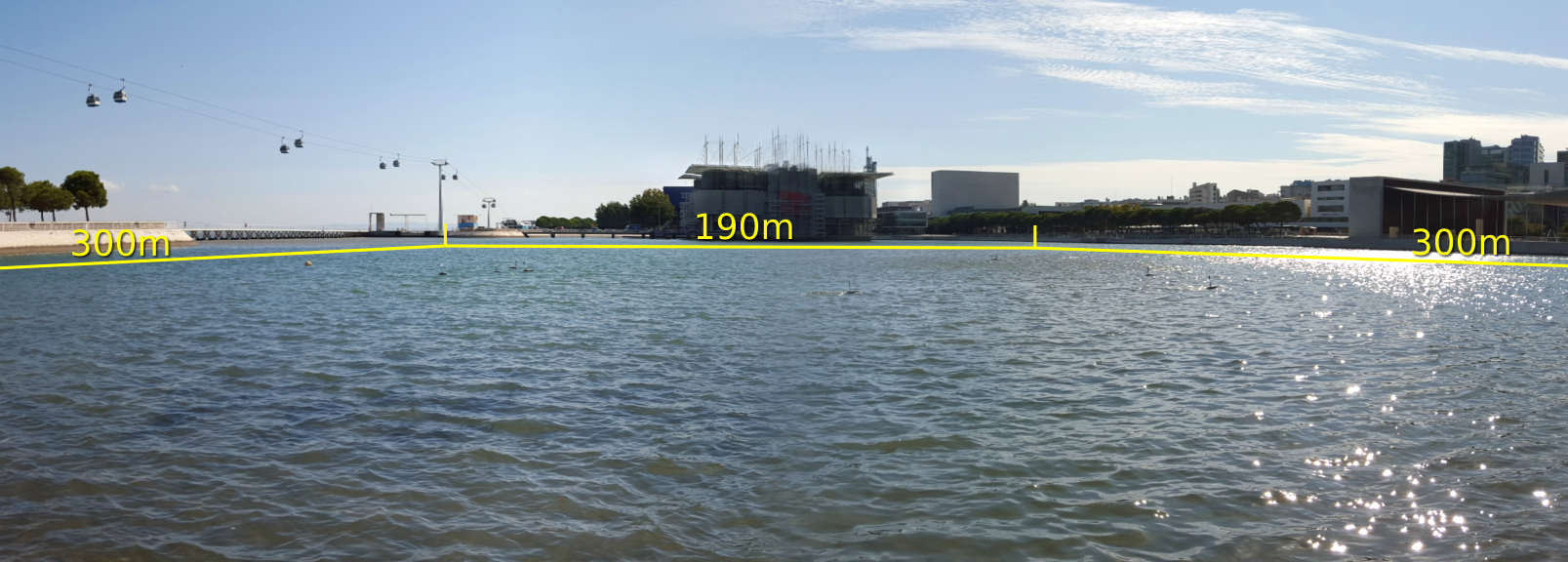}
	\caption{Panoramic photo of the location the experiments at Parque das Na\c{c}\~{o}es, Lisbon, Portugal.}
	\label{fig:expo}
\end{figure}

In this section, we assess the evolved controllers on real robots and compare their performance on real robots to their performance in simulation. The experiments reported in this section were conducted over the course of four days, at Parque das Na\c{c}\~{o}es, Lisbon, Portugal, in a semi-enclosed waterbody with an area of 300\,m $\times$ 190\,m which is connected to the Tagus river, see Fig.~\ref{fig:expo}. The wind speed ranged from 15 to 40\,km/h. The average drift speed of the robots was 0.2\,m/s, twice as high as we had used in simulation (see S1 Text). The experiments at Parque das Na\c{c}\~{o}es were carried out with the permission and cooperation of the nautical center Marina Parque das Nações, and preliminary experiments at the Lisbon Naval Base were carried out with the permission and cooperation of Arsenal do Alfeite and the Portuguese Navy. As described in Section~\ref{sec:methodology}, the top three controllers for each task were tested (referred to as Controllers 1, 2 and 3 in this section). Each controller was evaluated in three independent experiments (referred to as samples A, B and C). 

\subsection{Homing}
\label{sec:agg_wp}

\begin{figure}[h!]
	\centering
	\includegraphics[width=\textwidth]{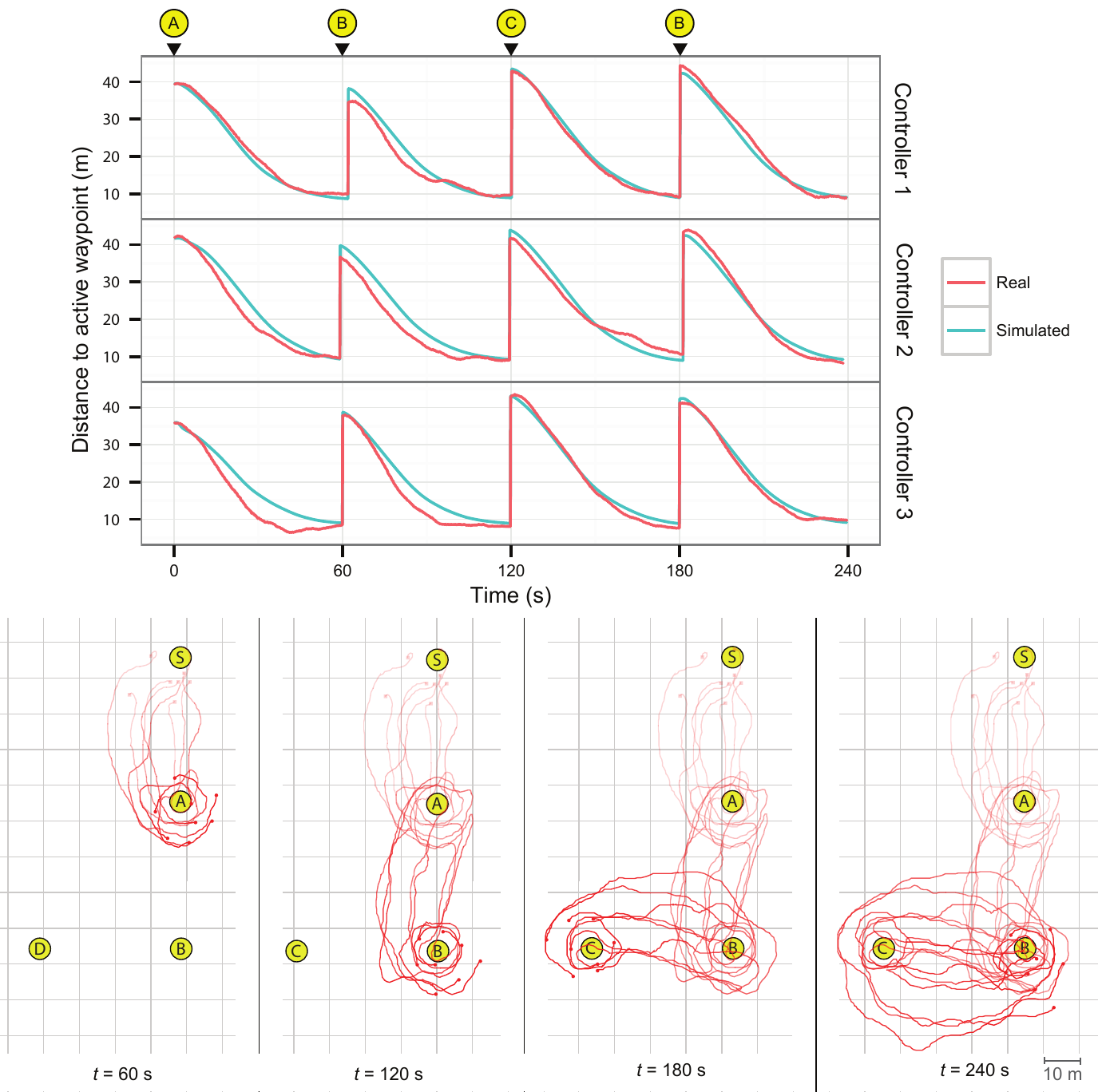}
	\caption{Real-world homing experiments with eight robots. The robots started around S. The active waypoint was then changed at 60 second intervals, in the order A$\to$B$\to$C$\to$B, for a total of four minutes per experiment. Top: comparison between the real and simulated robots, showing the average distance to the active waypoint, for similar conditions. The top of the figure shows the current active waypoint. Bottom: trajectories of the real robots for Controller 3. The waypoints are marked with yellow circles.}
	\label{fig:agg_wp_comparison}
\end{figure}

We setup a scenario where the swarm had to navigate sequentially to four waypoints to assess the performance of the evolved homing controllers. Each waypoint was placed at a distance of 40\,m from the previous one, and the active waypoint was changed periodically every 60\,s. Each controller was tested once in this scenario, for a total of 4 minutes. All three controllers tested in the real robots were able to navigate successfully to the waypoint while avoiding collisions with the neighboring robots. Fig.~\ref{fig:agg_wp_comparison} (top) shows the average distance to the active waypoint, compared to the same controller in the simulation environment. The controllers display a similar behavior and level of performance in reality and in simulation, arriving at the intermediate waypoints at approximately the same time, and keeping the same average distance to them (around 10\,m). Once the swarm arrived at one of the waypoints, the robots would start moving around the waypoint at slow speeds in concentric circles, see Fig.~\ref{fig:agg_wp_comparison} (bottom). 



\subsection{Dispersion}

In the real-robot dispersion tests, eight robots started in a cluster, and had to disperse to a distance of 20\,m from their nearest neighbor. The robots were initially placed randomly in a square-shaped area with a side-length of 28\,m, at a minimum distance of 5\,m from the nearest neighbor. Each experiment lasted for 90~seconds. 

\begin{figure}[h!]
\centering
\includegraphics[width=\textwidth]{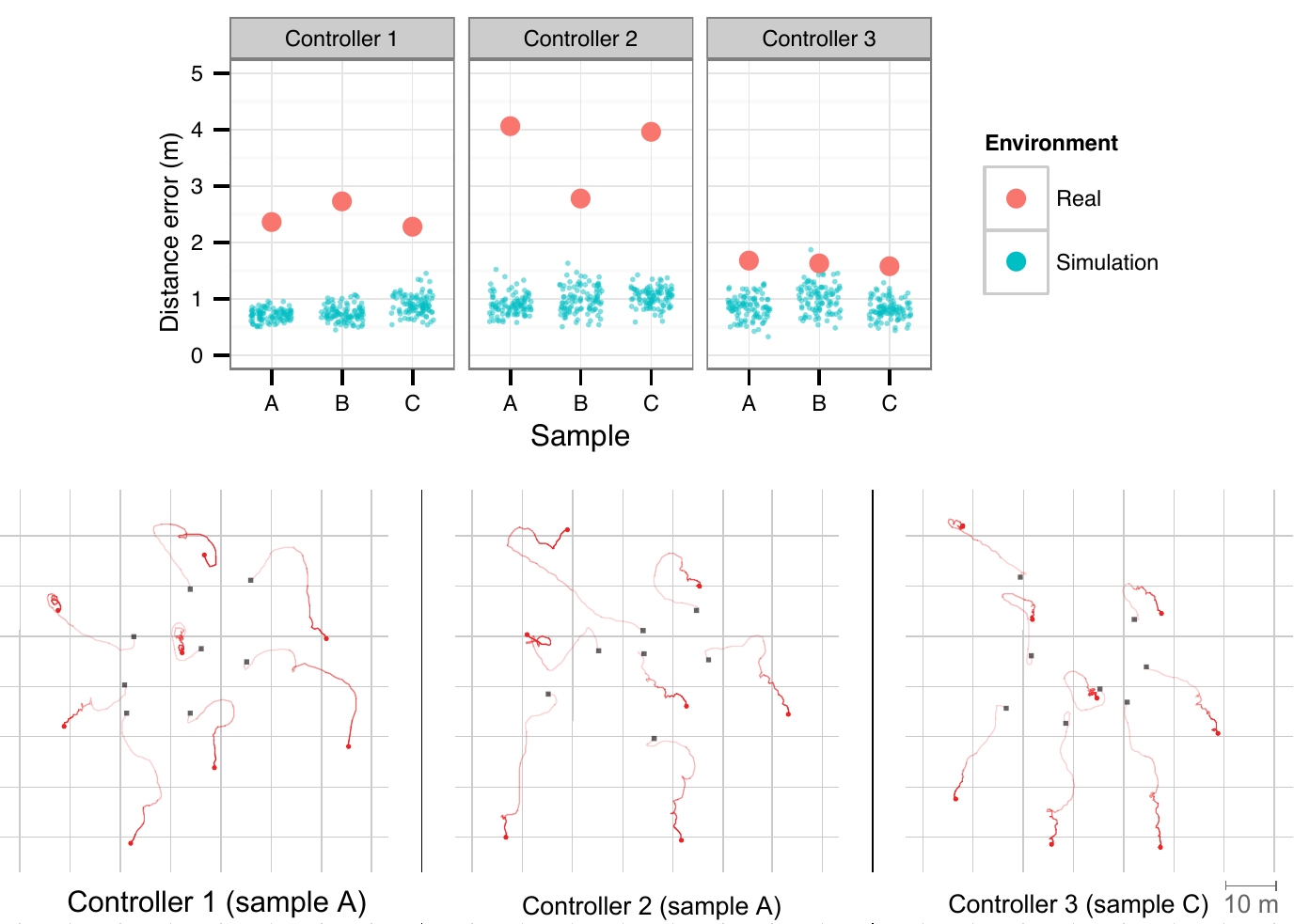}

\caption{Real-world dispersion experiments with eight robots, one for each controller tested, over a period of 90 seconds. Top: average error to target distance (20\,m) of the nearest robot in the last 10\,s of each dispersion experiment. Bottom: trajectories of the real robots. The black squares mark the starting positions, and the red circles mark the final positions.}
\label{fig:dispersion_comparison}
\end{figure}

The results in Fig.~\ref{fig:dispersion_comparison} show some differences in the performance of controllers on the real robots and in simulation. While performance was relatively homogeneous across all controllers in the simulation environment, with an error close to 1\,m to the target distance, no controller was able to achieve the same level of performance on real robots. Controller 3 achieved an average error of 2\,m on the real robots, but the other two controllers performed considerably worse, despite all of them displaying similar performance in simulation. Upon inspection of the behaviors, we observed that the robots in simulation relied on moving in circles at very low speeds to maintain their positions after dispersion. The same movement pattern was observed on the real robots. However, the circles performed by the real robots were larger, meaning that they tended to move away from their position, which is explained by the disparities between the motion model in simulation and the motion of the real robots.

\subsection{Clustering}

\begin{figure}[h!]
	\centering
    \includegraphics[width=\textwidth]{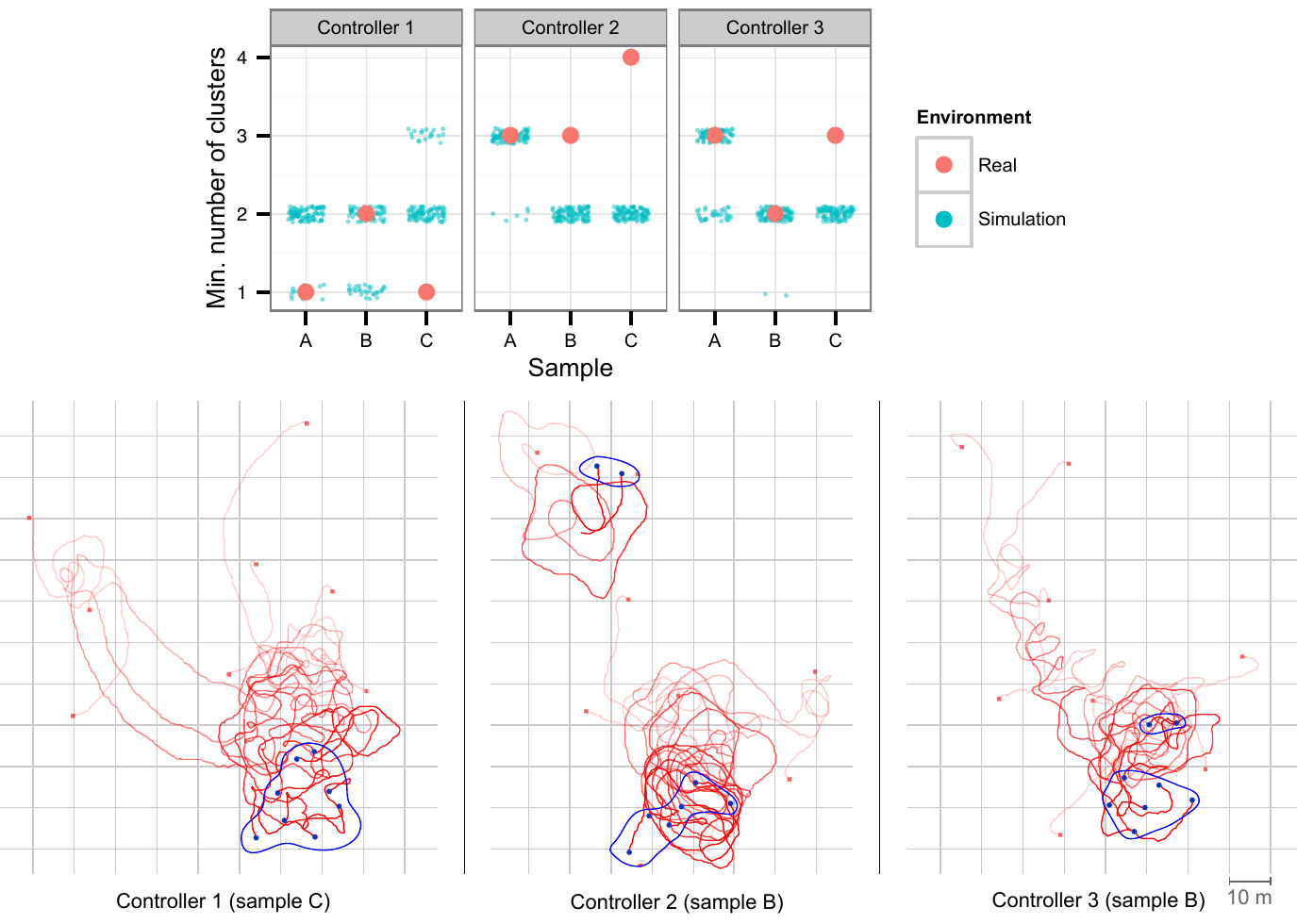}
	\caption{Real-world clustering experiments with eight robots, over a period of 180 seconds. Top: minimum number of clusters obtained in each sample. Bottom: trajectories of the real robots. The final clusters are highlighted in blue.}
	\label{fig:clustering_comparison}
\end{figure}

The clustering controllers were tested by randomly placing the real robots in an area of 100x100\,m, up to a maximum of 40\,m from the nearest robot. Each experiment lasted for 180~seconds. Results for the clustering experiments greatly varied depending on the initial conditions, but this variation was also observed in the simulation environment (see Fig.~\ref{fig:clustering_comparison}, top). If one or more robots were initially positioned far away from the remaining robots (and therefore not within communication range), the swarm would sometimes aggregate in more than one cluster, see example in Fig.~\ref{fig:clustering_comparison} (bottom, Controller 2). The results show that the controllers were generally able to cross the reality gap successfully. In the case of Controller 1, it had a better performance in the real-robot experiments than in simulation, by successfully aggregating into a single cluster in two of the three samples. Controller 2, on the other hand, performed worse than in simulation, and Controller 3 displayed a similar level of performance.


\subsection{Area Monitoring}

We assessed each of the three highest-performing area monitoring controllers in real robots in three areas: square, L-shaped, and rectangular. Each of the areas covered a total of 10,000\,m$^2$. It should be noted that the evolutionary process did not optimize the controllers for areas with these specific shapes, but rather for randomly generated shapes in order to promote the evolution of general behaviors (see Section~\ref{sec:evolutionary_setup}). Only the highest-performing controller was tested on real robots, since we did not observe any significant behavioral or performance differences in the other best controllers. Each experiment lasted for a total of five minutes.

\begin{figure}[h!]
	\centering
	\includegraphics[width=\textwidth]{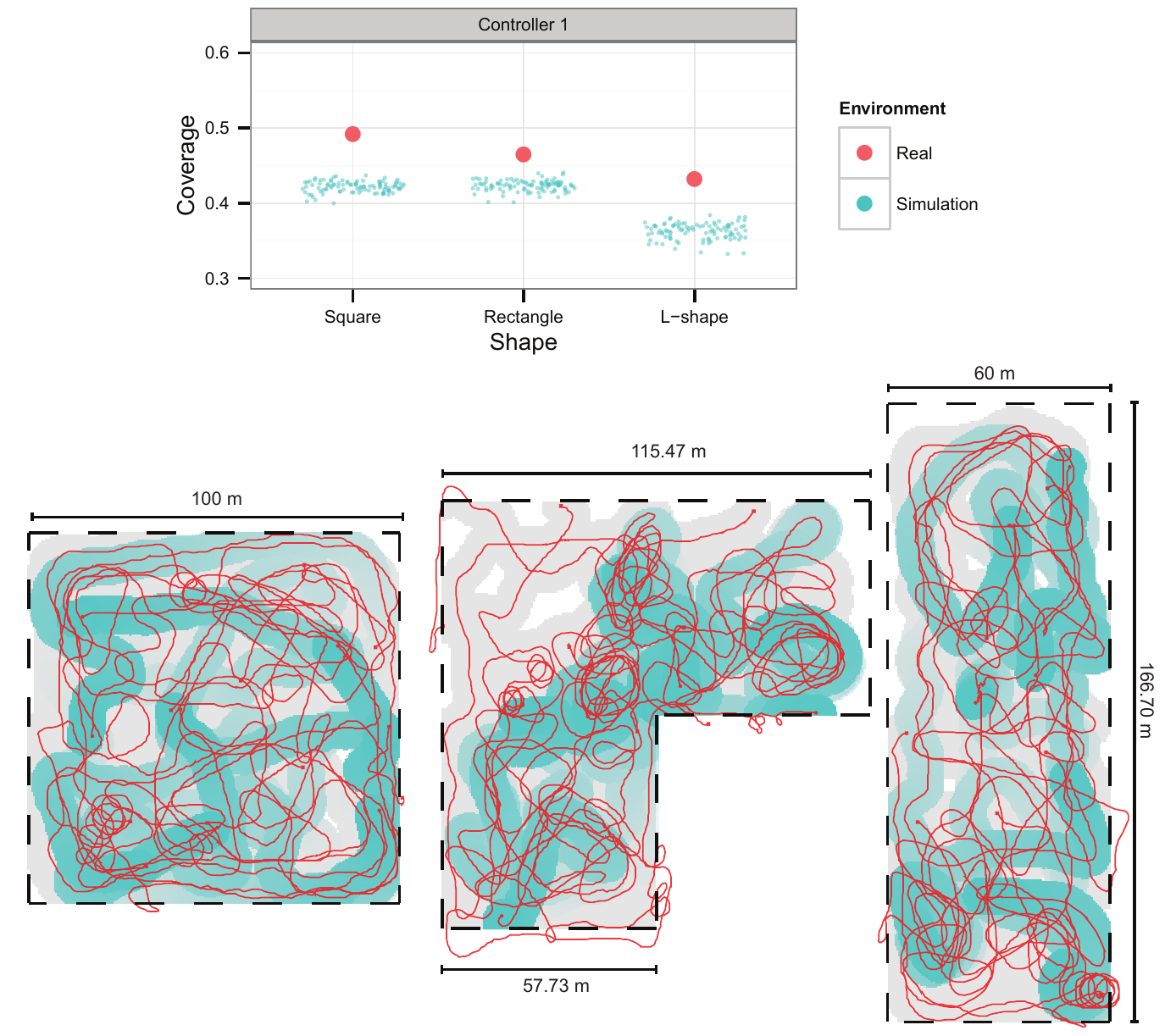}
	\caption{Real-world monitoring experiments with eight robots for Controller 1, over a period of five minutes. Top: coverage of the three different monitoring areas. Bottom: coverage maps in the experiments with the real swarm. The coverage of the area is presented in blue, and has a decay of 100\,s. Trajectories for the full duration of the task are presented in red, and all the areas visited by the robots are filled in gray.}
	\label{fig:monitor_comparison}
\end{figure}

The controllers successfully monitored the three different areas, performing significantly better on the real robots than in simulation (see Fig.~\ref{fig:monitor_comparison}, top). These results are explained by speed differences between the simulated robots and some of the real robots. The speed differences directly influence the coverage measure, since a faster robot will be able to cover a larger area in the same amount of time. In terms of behavior (Fig.~\ref{fig:monitor_comparison}, bottom), it is possible to see that the robots were able to effectively cover the square and rectangular areas, staying within the boundaries and moving away from the other nearby robots. The L-shaped area was also covered reasonably well, with the robots passing over all regions, but the intensity of the coverage was not uniform across the whole area. Although some robots would eventually move outside of the pre-defined boundaries (both in simulation and on the real robots) either because of wind, currents, GPS inaccuracies, or too many robots nearby, they would quickly return to the area inside the boundaries and resume monitoring.




\section{Scalability and Robustness}


Desirable characteristics of swarm behaviors include scalability and robustness~\cite{sahin2005}. Performance should increase as more units are added (until issues such as congestion and overcrowding become a limiting factor), and failures in individual units should not compromise the performance of the rest of the swarm. To evaluate if our swarm of real robots with evolved control would display these properties, we ran an additional series of experiments: (i)~scalability experiments for the dispersion and clustering tasks with four and six robots, (ii)~robustness experiments for the dispersion task where a second group of robots is added to the swarm after a period of time, and (iii)~robustness experiments for the monitoring task where robots are physically removed during the task executing (simulating faults), and then reintroduced after a period of time. All experiments were conducted in the real environment with the controllers that showed the highest performance in real hardware in the previous experiments.

\subsection{Scalability}
\label{sec:scalability}

\begin{figure}[h!]
\includegraphics[width=\textwidth]{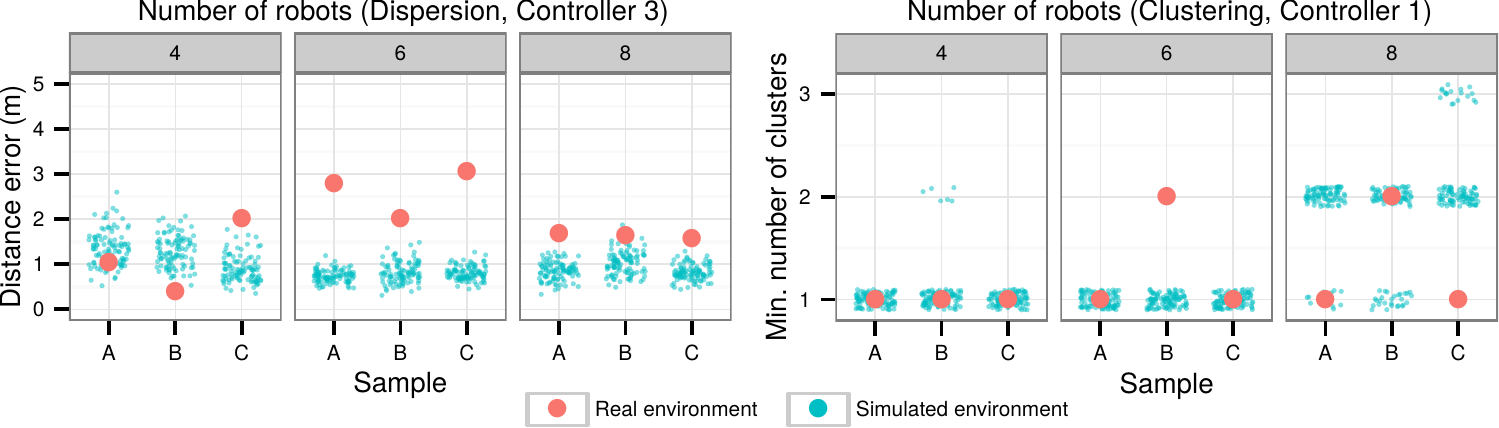}
\caption{Scalabiliy experiments with dispersion (Controller 3, left) and clustering (Controller 1, right) controllers. In each task, the same controller was used in a swarm of four, six, and eight robots, with three samples for each setup.}
\label{fig:scalability}
\end{figure}

The dispersion and clustering controllers were tested with four and six robots (in addition to the first experiments with eight robots) to determine if the evolved controllers were able to perform well independently of the number of robots in the swarm. Fig.~\ref{fig:scalability} summarizes the results for these experiments.

In the dispersion experiments, while the controllers were able to perform the task successfully with swarms of different sizes, the results show a slight performance decrease in the experiments with six robots. It must be noted, however, that the error obtained is still relatively low, when compared to the target distance of 20~m. In the clustering experiments, the swarm was able to aggregate into a single cluster in all three samples with four robots, and in two out of three samples with six and eight robots. Since the communication range is limited, subsets of robots may initially be outside communication range or lose contact during an experiment, which can result in two or more isolated clusters.

\subsection{Robustness}
\label{sec:robustness}

We setup a dispersion task that starts with a group of four robots ($G_a$). In the beginning of the experiment, $G_a$ start dispersing. After 60\,s, a group of four additional robots ($G_b$) starts moving towards the center of the swarm. When all robots of $G_b$ are at the center ($t=130$\,s), their presence disturbs the dispersion of $G_a$, forcing $G_a$ to separate even further. After the robots in $G_b$ have been in the center for 60\,s, they start the dispersion behavior along with the robots in $G_a$ (at $t=180$\,s).

\begin{figure}[h!]
\centering
\includegraphics[width=\textwidth]{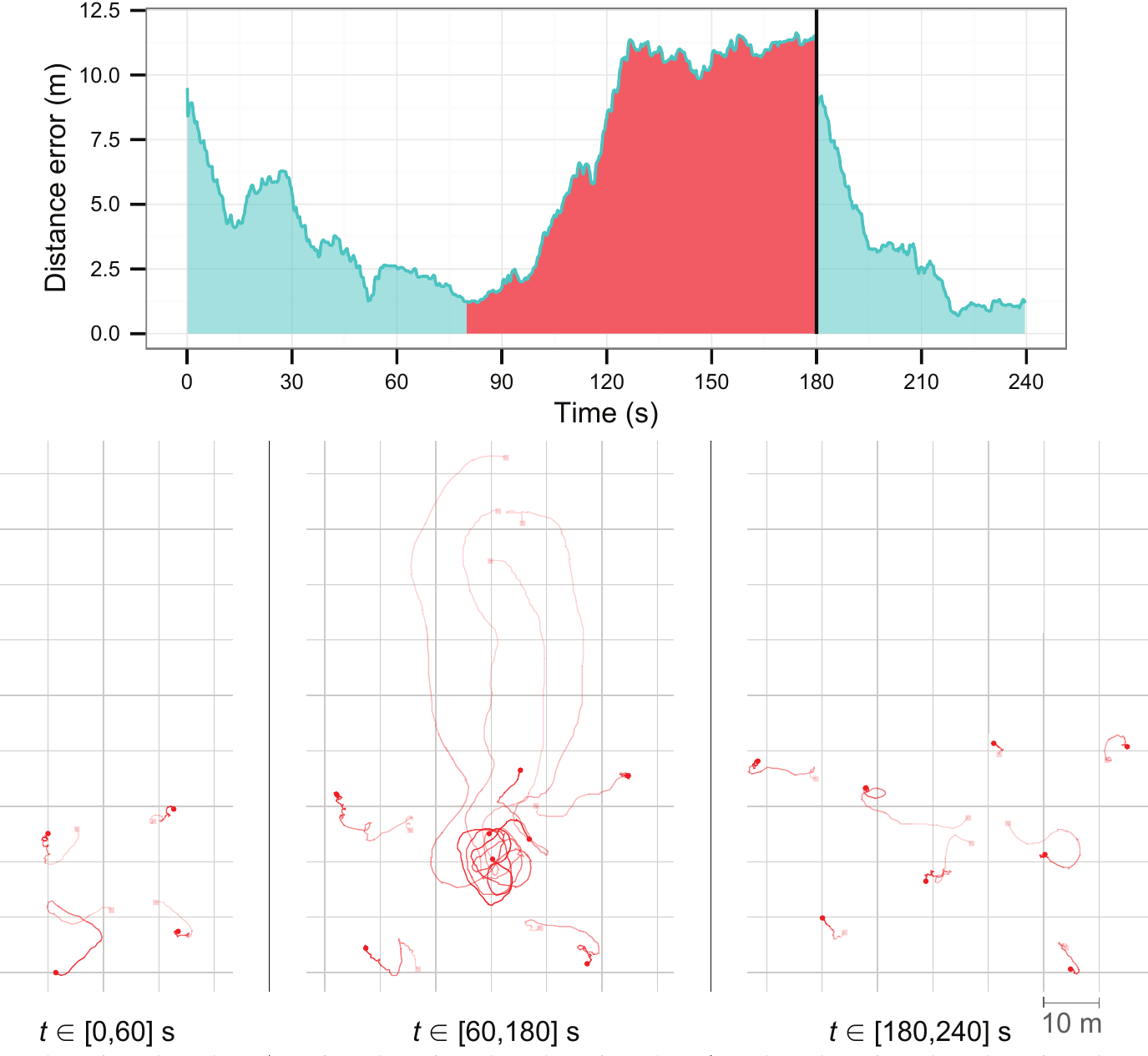}
\caption{Robustness experiments with Controller 3 of the dispersion behavior. The red area represents the period where the robots of $G_b$ are disturbing the dispersion of $G_a$, and the black vertical line at $t=180$\,s indicates the point where the robots in $G_b$ start dispersing, and where the distance error starts being measured for all eight robots.}
\label{fig:dispersion_adaptive}
\end{figure}

Fig.~\ref{fig:dispersion_adaptive} shows the results for the robustness experiments. When $G_b$ moves to the center of $G_a$, the robots in $G_a$ are forced to move away from one another since they try to stay at a fixed distance from the closest robot. When $G_b$ starts dispersing, robots from $G_a$ move further away to accommodate the movement pattern of $G_b$. When the whole swarm is dispersing, the distance error decreases with time as the robots continue adjusting their positions. The results demonstrate that the controllers were able to adjust to conditions that they were not exposed to during evolution, particularly the addition of new robots during task execution, and still carry out the task successfully.

\begin{figure}[h!]
\centering
\includegraphics{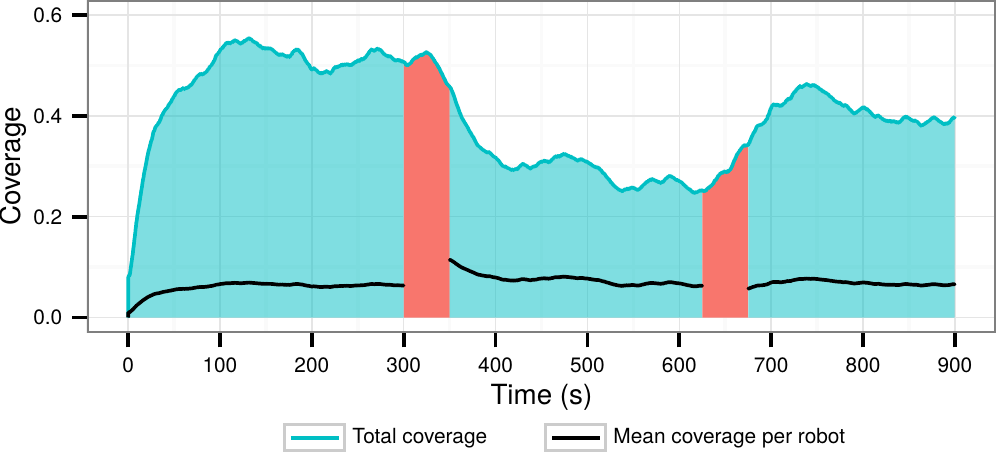}
\caption{Robustness experiments with Controller 1 of the monitoring behavior. The time regions highlighted in red correspond to the periods when robots where either entering or leaving the monitoring area.}
\label{fig:monitoring_adaptive}
\end{figure}

In an additional experiment, we physically removed and added units during the execution of the monitoring task. The area to monitor was a square with 100$\times$100\,m. The monitoring controller was executed for a total of 15~minutes, starting with eight robots. At the 5-minute mark, we removed four of the robots, and at the 10-minute mark, two new robots were added. Fig.~\ref{fig:monitoring_adaptive} shows the coverage of the monitoring area at any given time during the experiment. The performance decrease after the 5-minute (300~s) mark, and the consequent increase at the 10~minute (600~s) mark show that the performance of the evolved controllers tends to scale linearly with the number of robots executing the task (see mean coverage in Fig.~\ref{fig:monitoring_adaptive}). These results indicate that the evolved controller is robust to variations in the composition of the swarm during task execution. A video of the experiment can be found in the supplemental information (S2 Video).

\section{Multi-controller Mission}
\label{sec:multicontroller_mission}

In this section, we show how relatively simple swarm behaviors can be combined to produce more sophisticated behaviors that could be used in realistic applications. We integrate the highest-performing behaviors described in the previous section to accomplish a mission of sampling the water temperature over a given area of interest. The robots all started near a base station, outside the area of interest. The mission was decomposed in five sequential sub-tasks: (i) collectively navigate from the base station to the center of the area of interest (homing); (ii) disperse; (iii) monitor the area of interest; (iv) aggregate, using the clustering behavior; and finally (v) return to the base station (homing). The robots start collecting temperature data after reaching the center of the monitoring area, before sub-task (ii). The different behaviors were triggered sequentially to produce a complete mission, with each behavior lasting for a predefined amount of time.

\begin{figure}[h!]
\centering
\includegraphics[width=\textwidth]{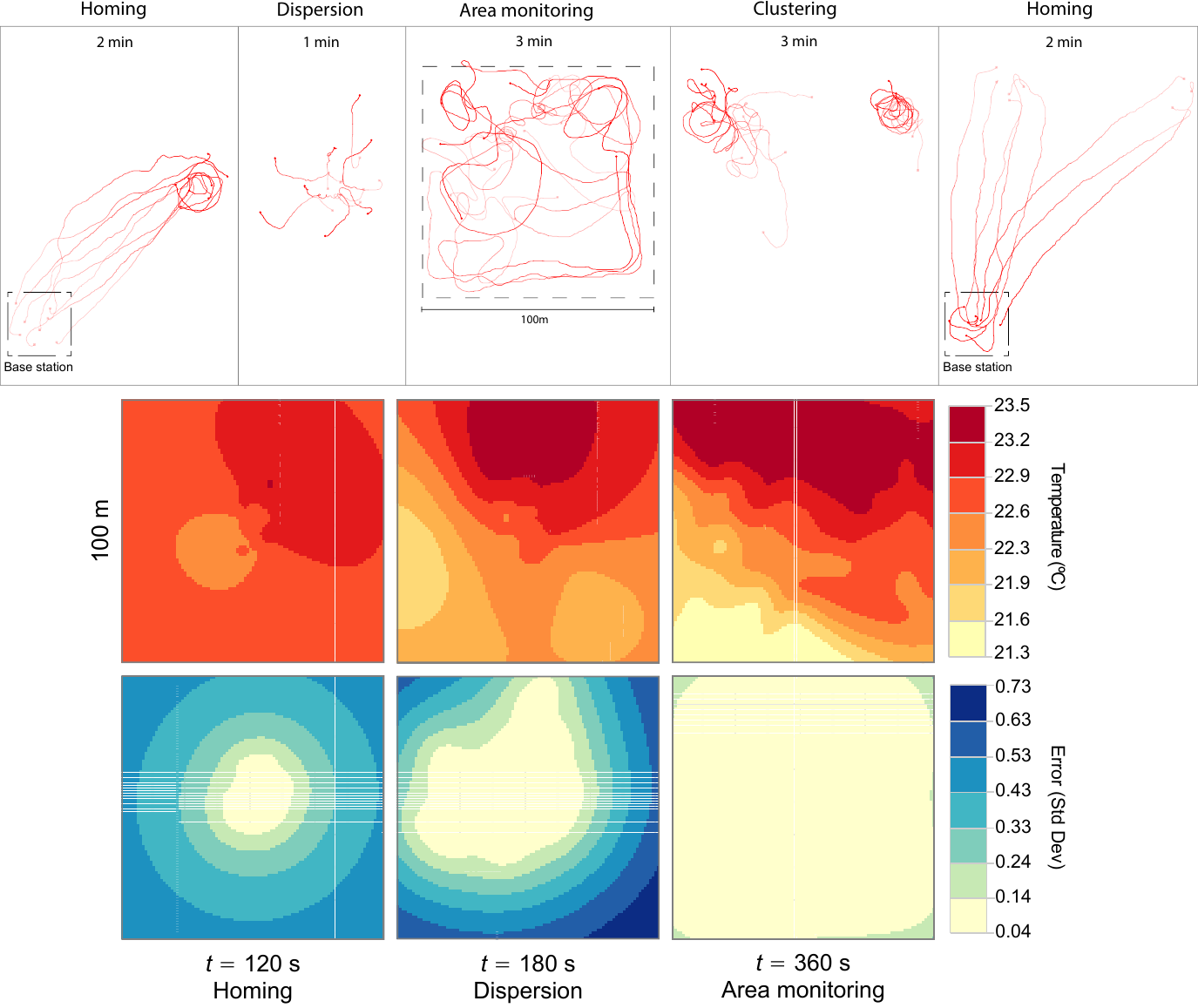}
\caption{Results for the multi-controller mission. Top: robot trajectories for each sub-task. Middle and bottom: temperatures in the monitoring area. Measurements taken by the robots' temperature sensors were spatially interpolated using kriging~\cite{stein2012interpolation}. Data collection started after the robots arrived at the waypoint ($t=100$\,s). The middle row shows the predicted temperatures, while the bottom row shows the estimated error of the predictions.}
\label{fig:composite}
\end{figure}

The trajectories of the real robots executing the multi-controller mission and the temperature data collected can be seen in Fig.~\ref{fig:composite}. The uncertainty of the temperature estimate deceases as the robots cover the area, resulting in a more accurate profile. At the end of the monitoring period ($t=360$\,s), the estimated standard deviation of the error was $<$ 0.14 for the whole area except in the corners where it was slightly higher ($<$ 0.24). A video of the experiment is available in S3 Video.



\section{Discussion}
\label{sec:discussion}

Our experiments showed that evolutionary computation is a viable approach for the synthesis of control for swarms of robots operating outside of controlled laboratory environments. Below, we discuss the main challenges we faced in evolving control for the SRS, and the properties observed in the SRS with evolved control.


\subsection{Evolutionary Process}
For the most part, the definition and configuration of the evolutionary process was straightforward. The main challenge in configuring the evolutionary algorithm was the definition of the fitness functions. The fact that we used common swarm robotics tasks facilitated the definition of the fitness functions, however, once we started considering more complex tasks (as the temperature monitoring task in Section~\ref{sec:multicontroller_mission}), it became hard to design effective fitness function that could express the task objectives. Moreover, previous works have shown that using standard evolutionary techniques for such complex and sequential tasks can be inadequate due to bootstrapping and deception issues~\cite{Nelson09fitnessfunctions,gomez1997incremental,mouret2008incremental,silva2015r}.


One solution is to synthesize the various \emph{building blocks} a-priori, and then combine them in order to solve complex tasks~\cite{duarte2015hybrid}. We have explored a variant of this line of research in this paper with the multi-controller mission (Section~\ref{sec:multicontroller_mission}), where a sequence of different behaviors were triggered to produce a complete mission, with each behavior executing for a predefined amount of time. However, as we have successfully shown in previous work, more complex decision schemes can be used to articulate different behaviors~\cite{duarte2014aquatic,duarte2014hierarchical,duarte2015hybrid}. In recent work~\cite{duarte2016}, we have taken the first steps towards the demonstration of hierarchical control in a real robotic swarm, and demonstrate how this control approach can produce modular and flexible behaviors.

\subsection{Transferability of Controllers}

One of the challenges of synthesizing control for aquatic surface robots is dealing with the unpredictable environmental conditions and unexpected circumstances associated with this environment. Many of the environmental variables affecting the robots can be extremely difficult, or even impossible, to model accurately. The simulation model was therefore a substantial simplification of reality. To compensate for the differences between the motion model and the real dynamics of the robots, as well as for other potential sources of stochasticity, we introduced a conservative amount of noise in the sensors, actuators, and environment, during evolution (Section~\ref{sec:methodology}). The introduction of noise is a computationally-effective way of promoting the evolution of robust controllers~\cite{Miglino96evolvingmobile}. In addition to noise, each controller was evaluated in multiple trials with variations in the task setup, such as the number of robots in each sample, or the initial positions of the robots and environmental features, which also contributed to the increase  of transferability of the evolved controllers.


Our experimental results showed that the evolved controllers were generally able to cross the reality gap. In a number of experiments, a speed difference between some of the real robots and the simulated robots (up to 20\%, in some cases) contributed to differences in terms of performance. While the behaviors of the real robots were generally similar to those observed in simulation, some motion patterns did not transfer well. With the dispersion controllers, for instance, in simulation the robots started moving in small circles to maintain their current position. In reality, however, these circles were larger, which resulted in a lower performance. This difference highlights the limitations of the simplified physics model we adopted in the simulation environment. We are considering two different approaches to address these issues in future work: (i)~further improve the simulated movement dynamics, thereby decreasing the mismatch between simulation and reality, and (ii)~select for controllers that rely on motion patterns that transfer well from simulation to reality, as proposed in previous works~\cite{koos2013transferability,cully2015robots,lehman2013encouraging}.

\subsection{Swarm Behaviors}

In this paper, we relied on four canonical swarm tasks (homing, dispersion, clustering, and area monitoring) to validate the basic properties of swarm robotics in real conditions. All the collective tasks presented in this paper can be classified as swarm robotics tasks, according to the classification proposed by Brambilla et al.~\cite{brambilla2013swarm}: (i)~robots are autonomous; (ii)~robots are situated in the environment; (iii)~robots’ sensing and communication capabilities are local (up to 40\,m, while the robots can be over 200\,m apart from each other); (iv)~robots do not have access to centralized control and/or to global information; and (v)~robots cooperate to solve a given task. Additionally, {\c{S}}ahin~\citep{sahin2005} states three properties as motivations for swarm robotics, which we were able to verify with the experiments described in the paper:

\vspace{-.5em}\paragraph{Robustness:} We demonstrated robustness explicitly with the monitoring task, by removing robots during task execution (see Section~\ref{sec:robustness}). In all tasks, there were also sporadic hardware failures, which resulted in individual failures that did not impact the behavior of the group (Section~\ref{sec:robustness}). 

\vspace{-.5em}\paragraph{Flexibility:} The flexibility of the swarm could be observed in the different coordination strategies that were adopted by the swarm, when using controllers for different tasks. Furthermore, we showed the behavioral modularity of the system by assembling a high-level controller that relied on the combination of multiple previously evolved controllers (see Section~\ref{sec:multicontroller_mission}).

\vspace{-.5em}\paragraph{Scalability:} We studied scalability explicitly using the clustering and dispersion behaviors (Section~\ref{sec:scalability}). We tested the same controller with different swarm sizes (from four to eight robots), and the performance and behavior of the swarm was similar regardless of the swarm size. In the monitoring task, we showed that the performance of the swarm (amount of space covered) was proportional to the number of robots (Section~\ref{sec:robustness}).


\subsection{Robustness to Hardware Faults}
\label{sec:robustness_to_hardware_faults}

The use of robots in real-world conditions will often imply hardware malfunctions, caused by prolonged use and environmental factors. In our experiments, the main source of faults were the robots' motors, which sporadically and temporarily stopped, causing the robot to stop completely or start moving in circles. Other problems included the decrease of motor power with lower battery levels, and often erroneous GPS readings. These faults occurred often during the execution of all experiments reported in this paper. Even though these faults were not contemplated in simulation, and therefore were not taken into account during the evolutionary process, they did not compromise the overall performance of the swarm, highlighting the inherent robustness of swarm behaviors. In future work, we will study the fault-tolerance properties of the system more systematically by injecting different types of faults~\cite{christensen2008fault}, and by measuring the impact of these faults in the performance of the swarm. 

An additional source of uncertainty, related to the hardware, was the heterogeneity among the robots of the swarm. Robots within the swarm often had different battery levels and motors with slight performance differences. We measured this heterogeneity in the monitoring task, where all robots display similar behaviors. In the real swarm, the difference between the slowest and the fastest robot was up to 0.6\,m/s at certain times, while for the same task, in simulation, the difference between the slowest and the fastest robot was just 0.1\,m/s. This suggests that the real swarm had to deal with a much higher degree of heterogeneity than what was contemplated during the evolutionary process.

\subsection{Towards Real World Applications}

In this study, we have demonstrated that swarm robotics systems can effectively operate in real-world conditions. There are, however, a number of control-related issues to be addressed before the practical use of swarms of robots in real-world applications. Prominent challenges include:

\vspace{-.5em}\paragraph{Integration of mission-specific sensors:} In the experiments presented in this paper, the robot controllers relied mostly on their position and on the position of the neighboring robots for the behavioral decision process. The integration of data gathered from onboard sensors such as cameras or sonars, may however be an essential input for the controllers in other tasks. The challenge of evolving controllers capable of dealing with such data is twofold: (i)~the data might not be trivial to decompose into meaningful information that can be fed to neural controllers; and (ii)~the sensors would have to be accurately modeled in simulation.

\vspace{-.5em}\paragraph{Robustness to faults:} Another significant challenge is endowing controllers with the capability to cope with faults within the swarm. Although our experiments confirmed that swarm behaviors inherently have some degree of tolerance for individual faults, in real-world applications it may be necessary to guarantee that faulty individuals will not compromise the whole swarm. Approaches for fault detection~\cite{christensen2008fault}, fault tolerance~\cite{tarapore2015err}, and online adaptation~\cite{cully2015robots} must therefore be implemented in the system.

\vspace{-.5em}\paragraph{Safety:} The robots should have the capability of dealing with external factors, such as nearby vessels, in a way that guarantees the safety of all parties and conforms with the existing regulations. Combining such fail-safe behaviors with the mission controllers may require the combination of evolved and manually programmed behaviors~\cite{duarte2014hierarchical}.

\vspace{-.5em}\paragraph{Behavior flexibility and modularity:} One concern when using evolutionary robotics is the difficulty in rapidly producing new behaviors or modify existing ones. One possible solution for this problem is to compose behaviors for new tasks based on existing building blocks~\cite{duarte2015hybrid}. In this study, we showed how multiple evolved behaviors could be sequentially arranged in order to perform an environmental monitoring task. Future work should explore the usage of more capable techniques for combining controllers, such as hierarchical composition~\cite{duarte2015hybrid,duarte2016}.

\section{Conclusion}

In this paper, we present for the first time a swarm robotics system with evolved control operating in a real-world uncontrolled environment. Our experiments relied on a swarm of small and inexpensive aquatic surface robots that used distributed self-organized control and local communication. In a simulation environment, we evolved controllers for four common collective tasks: homing, clustering, dispersion, and area monitoring. The evolved controllers were then systematically evaluated in the real robots in a large and uncontrolled aquatic environment. Our results showed that the controllers displayed similar behaviors and levels of performance in the real swarm as those observed in simulation. Moreover, we verified key desirable properties of swarm robotics systems, namely robustness, flexibility, and scalability. We finally showed how simple evolved behaviors can be combined to produce a complete mission, by carrying out a complete environmental monitoring task.

Past studies in the field of swarm robotics and evolutionary robotics have been confined either to simulation (the vast majority), or to highly-controlled laboratory conditions. Our study contrasts with the previous work in these fields, as it relies on an evolved swarm of robots operating in a realistic and uncontrolled environment. Overall, our research demonstrates the potential of both swarm robotics systems and evolutionary robotics techniques as viable approaches for controlling multirobot systems in real-world scenarios.

\section*{Supporting Information}

\paragraph{S1 Text. Experimental Details.} Document detailing the aquatic robotic platform and the experimental parameters of both the simulated and real experiments, in some cases repeating and extending the information given in the main manuscript. (PDF)

\paragraph{S2 Video. Adaptive monitoring experiment.} Traces of the robots' positions when performing the experiment where the robustness of the swarm is assessed using the area monitoring task, as described in Section~\ref{sec:robustness}.

\paragraph{S3 Video. Multi-controller mission.} Video of the experiment detailed in Section~\ref{sec:multicontroller_mission}, where the different swarm behaviors are composed sequentially in order to produce control for a temperature-sampling mission.

\bibliographystyle{plos2015}



\section*{Supplementary material}

# Evolution of Collective Behaviors for a Real Swarm of Aquatic Surface Robots — Supporting Information

Miguel Duarte[1,2,3,*], Vasco Costa[1,2,3], Jorge Gomes[1,2,4], Tiago Rodrigues[1,2,3], Fernando Silva[1,2,4], Sancho Moura Oliveira[1,2,3], Anders Lyhne Christensen[1,2,3]

1 BioMachines Lab, Lisbon, Portugal
2 Instituto de Telecomunicações, Lisbon, Portugal
3 University Institute of Lisbon (ISCTE-IUL), Lisbon, Portugal
4 BioISI, Faculdade de Ciências, Lisbon, Portugal

* miguel_duarte@iscte.pt

## Experimental Details

### Robotic Platform

We developed and produced a total of 10 relatively small (60 cm) and inexpensive ($\approx$ 300 EUR/unit) robots. We used digital manufacturing techniques to produce the robots, such as fused deposition 3D printing and CNC milling. Furthermore, we used widely available and off-the-shelf hardware in order to keep costs low, see Table 1 and Fig. 1. The robot is a differential drive monohull boat and its physical and dynamic properties are presented in Table 2. The Raspberry Pi 2 single-board computer was used for the control unit of each robot, and communication is achieved using an ad-hoc wireless network. A Kalman filter was applied to the GPS and compass readings of the real robots before they are used to compute sensory readings for the controller. Schematics, 3D models, and source code are available at `http://biomachineslab.com/aquaticdrone`.

Robots communicate with neighboring robots and with a base station using a IEEE 802.11g based ad-hoc wireless network (Wi-Fi). In order to assess the range of the chosen wireless adapter, we conducted empirical tests with the robots floating on the water surface, and achieved communication up to 40 m. When the swarm of robots is deployed, inter-robot communication is achieved by broadcasting messages. Each robot transmits a short status message, indicating its identification, position, and orientation. The status message is broadcast every second allowing neighboring robots to sense one another.

In order to monitor the swarm, an application monitored the messages that were sent by the robots in the swarm and displayed the locations in a map. To increase the range at which the inter-robot communication could be eavesdropped, an Ubiquiti BULLET-M2-HP was used at the base station, effectively increasing the monitoring range up to 300 m. The application was also used to send commands and messages to the robots. Examples include starting or stopping a specific controller, sending a list of waypoints or a geo-fence, and updating the onboard control software.

### Experimental Parameters

Table 3 lists the parameters used in our experiments, both in the simulation environment and in the real experiments. The noise that was applied in simulation during evolution is described below. All random numbers were drawn from uniform distributions. Regarding the movement model of the simulated robot, the values were taken by systematically performing tests with the real robot at different speeds and headings. The simulated dynamics were implemented taking into account the measurements obtained from these tests, and match the physical properties described in Table 2.

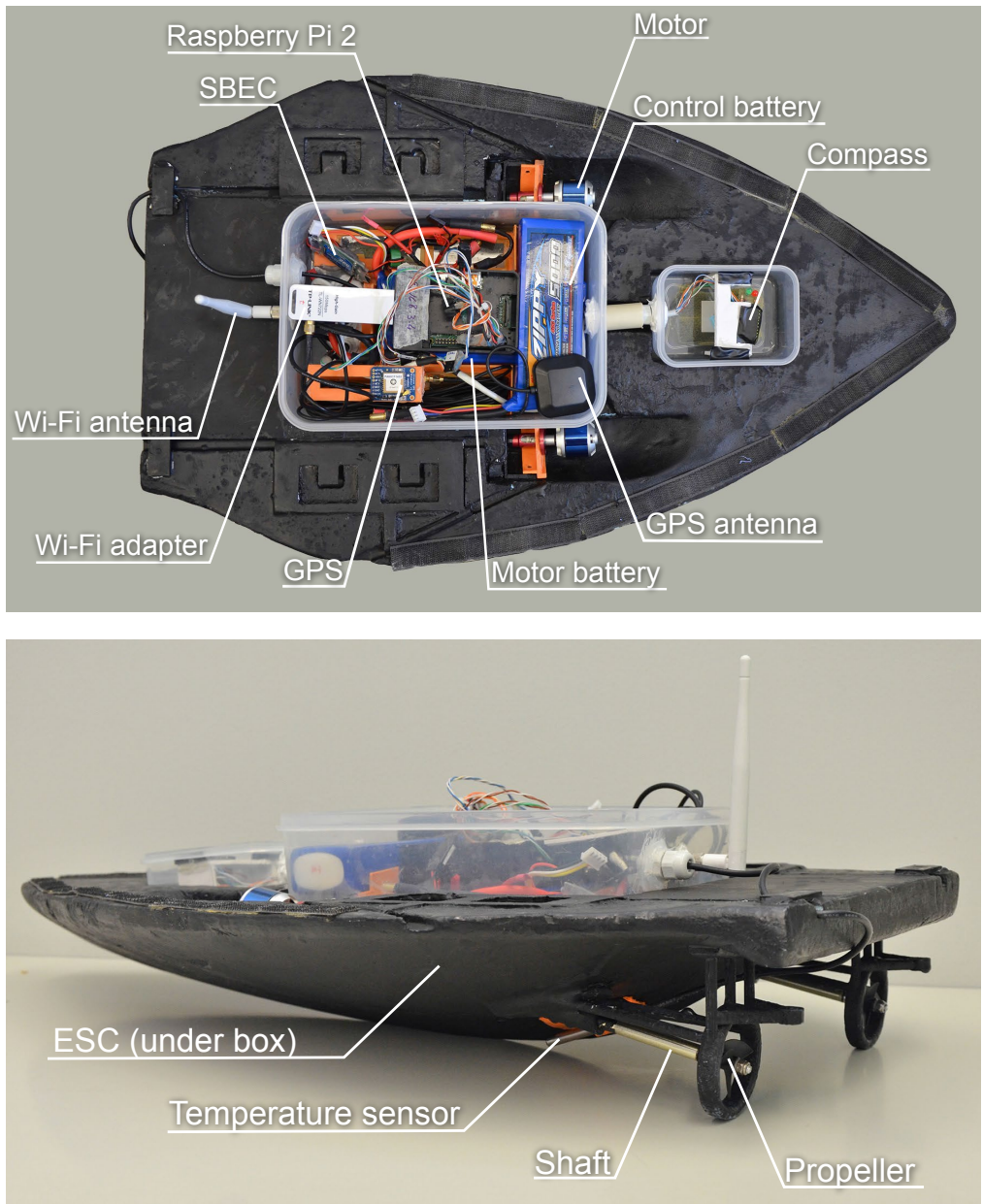

**Figure 1.** Top and side view of the final robot prototype with a description of the components.

- **GPS noise:** upper limit for noise added to the robots' GPS unit at every simulation timestep. The value was taken from the technical specification of the GPS used in our robots.

- **Compass noise:** upper limit for noise added to the robot's compass at every simulation timestep, chosen from empirical tests with the LSM303D compass.

- **Motor delay:** fixed delay between executing a motor speed command and observing a reaction in the movement of the robot.

- **Heading offset:** upper limit for noise added to the heading of the robot. The value is

**Table 1.** Robotic platform component list.

| Component | Make & Model |
|---|---|
| Motors (A) | NTM Prop Drive Series 28-30 A 750 kv / 140w |
| Motors (B) | Emax 2215/25 950 kv 2-3S |
| Shaft | 4 mm drive shaft |
| Shaft sleeve | 255 mm boat shaft sleeve |
| Propellers | 3-blade 28 mm |
| ESC | HobbyKing 50 A Boat ESC |
| Control battery | Zippy 40C Series 5000 mA 3 S LiPo |
| Motor battery | Zippy 30C Series 8000 mA 3 S LiPo |
| GPS | Adafruit Ultimate GPS Breakout |
| Compass | STMicroelectronics LSM303D |
| Water temperature sensor | DS18B20 |
| Onboard computer | Raspberry Pi 2 |
| Wi-Fi adapter | TP-Link TL-WN722N |
| Hull material | Extruded Polystyrene (XPS) + fiberglass with epoxy resin |
| Structural components | 3D printed Polylactic Acid (PLA) |
| Electronics enclosure | 2.5 L watertight plastic box |
| Compass enclosure | 0.4 L watertight plastic box |

**Table 2.** Measured movement dynamics and physical properties.

| Parameter | Value | Parameter | Value |
|---|---|---|---|
| Size (L × W × H) | 65 × 40 × 15 cm | Weight | 3 Kg |
| Minimum speed | 0.3 m/s | Maximum speed | 1.7 m/s |
| Maximum turning radius | 90 °/s | Maximum acceleration | 1.7 m/s$^2$ |
| Time from full speed to stop | 5 s | Autonomy (at full speed) | 1h30m |

set individually for each robot at the beginning of a sample.

**Speed offset:** upper limit for noise added to the speed of the robot. The value is set individually for each robot at the beginning of a sample.

**Motor output noise:** upper limit for noise added to the output of the controllers at every simulation step.

**Drift speed:** upper limit for the translation component added to all robots. The drift is intended to simulate the effects of water currents and wind. The value is chosen at the beginning of each sample, along with a random orientation, and is added to the position of all robots at every simulation timestep.

**Table 3.** Parameters used in the experiments.

| Parameter | Value | Parameter | Value |
|---|---|---|---|
| **NEAT** | | | |
| Population size | 150 | Target species count | 5 |
| Recurrency allowed | true | Mutation prob. | 25% |
| Prob. add node | 3% | Prob. mutate bias | 30% |
| Prob. add link | 5% | Crossover prob. | 20% |
| **Simulation noise** | | | |
| GPS noise | 1.8 m | Compass noise | 10° |
| Motor delay | 500 ms | Heading offset | 5% |
| Speed offset | 10% | Motor output noise | 5% |
| Drift speed | [0,0.1] m/s | | |
| **Homing task** | | | |
| Generations | 100 | Trial length (evolution) | 100 s |
| Deploy area (evo.) | 50 m × 50 m | Waypoint distance[1] (evo.) | [0,50] m |
| Robot sensor range | 20 m | Waypoint sensor range | 10 m |
| Initial robot separation | > 5 m | Trial length (real) | 240 s |
| Number of waypoints[2] (real) | 4 | Distance between WPs (real) | 40 m |
| **Dispersion task** | | | |
| Generations | 100 | Trial length (evo.) | 100 s |
| Deploy area (evo.) | 30 m × 30 m | Target distance[3] | 20 m |
| Robot sensor range | 40 m | Initial robot separation | > 5 m |
| Trial length (real) | 90 s | Deploy area, 4 robots (real) | 20 m × 20 m |
| Deploy area, 6 robots (real) | 24 m × 24 m | Deploy area, 8 robots (real) | 28 m × 28 m |
| **Clustering task** | | | |
| Generations | 400 | Trial length (evo.) | 200 s |
| Initial robot separation | [20,40] m | Robot sensor range | 40 m |
| Deploy area | 100 m × 100 m | Clustering threshold[4] | 7 m |
| Trial length (real) | 180 s | | |
| **Area monitoring task** | | | |
| Generations | 100 | Trial length (evo.) | 200 s |
| Robot sensor range | 40 m | Geo-fence sensor range | 40 m |
| Monitoring area (evo.) | [0.5,1.7] ha | Deploy area | 1.44 ha |
| Initial robot separation | > 5 m | Trial length (real) | 300 s |
| Monitoring area (real) | 1 ha | Deploy area (real) | 1 ha |

[1] Distance from the center of the deploy area.
[2] In the real experiments, the robots had to navigate through a sequence of waypoints.
[3] Distance that the robots should maintain from each other.
[4] Maximum distance for two robots to be considered as part of the same cluster.




\end{document}